\documentclass[pdflatex,sn-apa]{sn-jnl}
\usepackage[ruled,vlined,algo2e]{algorithm2e}
\usepackage[caption=false,font=footnotesize]{subfig}
\usepackage{siunitx}
\usepackage{enumitem}
\usepackage{bm}
\usepackage{xfrac}
\usepackage[capitalize]{cleveref}
\usepackage{xcolor}

\jyear{2021}%

\renewcommand{\set}[1]{\mathcal{#1}}
\newcommand{\mat}[1]{\bm{#1}}
\renewcommand{\vec}[1]{\mat{#1}}

\newcommand{\given}{\colon}
\renewcommand{\R}{\mathbb{R}}
\newcommand{\norm}[1]{\left\|#1\right\|}
\newcommand{\abs}[1]{\left\lvert#1\right\rvert}

\DeclareMathOperator*{\argmin}{arg\,min}

\newcommand{\ltg}{{\textit{local2global}}}

\newcommand{\bb}{\hspace{-1mm} $\bullet$ }
\begin{document}
\renewcommand{\doi}[1]{\url{https://doi.org/#1}}

\title[Local2Global]{Local2Global: A distributed approach for scaling representation learning on graphs}


\author*[1]{\fnm{Lucas G. S.} \sur{Jeub}}\email{lucasjeub@gmail.com}

\author[1,2]{\fnm{Giovanni} \sur{Colavizza}}\email{g.colavizza@uva.nl}

\author[3]{\fnm{Xiaowen} \sur{Dong}}\email{xdong@robots.ox.ac.uk}

\author[1,4]{\fnm{Marya} \sur{Bazzi}}\email{marya.bazzi@warwick.ac.uk}

\author[1,5]{\fnm{Mihai}\sur{Cucuringu}}\email{mihai.cucuringu@stats.ox.ac.uk}

\affil[1]{\orgname{The Alan Turing Institute}, \orgaddress{\street{96 Euston Road}, \city{London}, \postcode{NW1 2DB}, \country{United Kingdom}}}

\affil[2]{\orgdiv{Institute for Logic, Language and Computation (ILLC)}, \orgname{University of Amsterdam}, \orgaddress{\street{Science Park 107}, \city{Amsterdam}, \postcode{1098 XG}, \country{Netherlands}}}

\affil[3]{\orgdiv{Department of Engineering Science}, \orgname{University of Oxford}, \orgaddress{\street{Eagle House, Walton Well Road}, \city{Oxford}, \postcode{OX2 6ED}, \country{United Kingdom}}}

\affil[4]{\orgdiv{Department}, \orgname{University of Warwick}, \orgaddress{\street{Street}, \city{City}, \postcode{610101}, \state{State}, \country{Country}}}

\affil[5]{\orgdiv{Department of Statistics \&  Mathematical Institute}, \orgname{University of Oxford}, \orgaddress{\street{24-29 St Giles'}, \city{Oxford}, \postcode{OX1 3LB}, \country{United Kingdom}}}

\abstract{We propose a decentralised ``{\ltg}'' approach to graph representation learning, that one can a-priori use to scale any embedding technique.
Our {\ltg} approach proceeds by first dividing the input graph into overlapping subgraphs (or ``\textit{patches}'') and training local representations for each patch independently.
In a second step, we combine the local representations into a globally consistent representation by estimating the set of rigid motions that best align the local representations using information from the patch overlaps, via group synchronization.
A key distinguishing feature of {\ltg} relative to existing work is that patches are trained independently without the need for the often costly parameter synchronization during distributed training.
This allows {\ltg} to scale to large-scale industrial applications, where the input graph may not even fit into memory and may be stored in a distributed manner. We apply {\ltg} on data sets of different sizes and show that our approach achieves a good trade-off between scale and accuracy on edge reconstruction and semi-supervised classification. We also consider the downstream task of anomaly detection and show how one can use {\ltg} to highlight anomalies in cybersecurity networks.
}

\keywords{scalable graph embedding, distributed training, group synchronization}



\maketitle

\section{Introduction}
\label{sec:introduction}

The application of deep learning on graphs, or Graph Neural Networks (GNNs), has recently become popular due to their ability to perform well on a variety of tasks on non-Euclidean graph data.
Common tasks for graph-based learning include link prediction (i.e., predicting missing edges based on observed edges and node features), semi-supervised node classification (i.e., classifying nodes based on a limited set of labeled nodes, node features, and edges), and unsupervised representation learning (i.e., embedding nodes in a low-dimensional space $\R^d$, with $d$ much smaller than the number of nodes in the graph).
Among the significant open challenges in this area of research is the question of scalability both during training and inference. 
Cornerstone techniques such as Graph Convolutional Networks (GCNs) \citep{Kipf2017} make the training dependent on the neighborhood of any given node. Since in many real-world graphs the number of neighbors grows exponentially with the number of hops taken, the scalability of such methods is a significant challenge. In recent years, several techniques have been proposed to render GNNs more scalable, including layer-wise sampling \citep{Hamilton2018} and subgraph sampling \citep{Chiang2019} approaches (see \cref{sec:related}).

We contribute to this line of work by proposing a decentralized divide-and-conquer approach to improve the scalability of network embedding techniques. While our work focuses on node embeddings, our proposed methodology can be naturally extended to edge embeddings.  
Our ``{\ltg}'' approach proceeds by first dividing the network into overlapping subgraphs (or ``patches'') and training separate local node embeddings for each patch (local in the sense that each patch is embedded into its own local coordinate system).
The resulting local patch node embeddings are then transformed into a global node embedding (i.e.\ all nodes embedded into a single global coordinate system) by estimating a rigid motion applied to each patch using the As-Synchronized-As-Possible (ASAP) algorithm \citep{Cucuringu2012a,Cucuringu2012b}.
A key distinguishing feature of this ``decentralised'' approach is that we can train the different patch embeddings  separately  and   
independently,  without the need to keep parameters synchronised.
The benefit of {\ltg} is fourfold: 
\bb (1) it is highly parallelisable as each patch is trained independently;
\bb (2) it can be used in privacy-preserving applications and federated learning
setups, where frequent communication between devices is often a limiting factor \citep{Kairouz2021},
or ``decentralized'' organizations, where one needs to simultaneously consider data sets from different   departments;
\bb (3) it can reflect varying 
structure across a graph through asynchronous parameter learning; and 
\bb (4) it is a generic approach that, in contrast to most existing approaches, can be applied to a large variety of embedding techniques. 

We assess the performance of the resulting embeddings on three tasks: one edge-level task, i.e., edge reconstruction, and two node-level tasks, i.e., semi-supervised classification and anomaly detection. In the edge reconstruction task, we use the learned node embeddings to reconstruct known edges in the network, verifying that the embedding captures the network information.
In semi-supervised node classification, one has a few labeled examples and a large amount of unlabeled data, and the goal is to classify the unlabeled data based on the few labeled examples and any structure of the unlabeled data.
Here we propose to first train an embedding for the network in an unsupervised manner and then train a classifier on the embedding to predict the labels.
Finally, as an application we have applied our method to the problem of anomaly detection where a low-dimensional representation that denoises and preserves the intrinsic structural properties of the data is essential.

The remainder of this paper is structured as follows. We give a detailed overview of related work in \cref{sec:related} and describe our methodology in \cref{sec:l2g}. We show and interpret the results of numerical experiments on a variety of tasks and data sets in \cref{sec:semi-supervised} and \cref{sec:anomaly}. We conclude and offer directions for future work in \cref{sec:conc}. Our reference implementations of the {\ltg} patch alignment method\footnote{\url{https://github.com/LJeub/Local2Global}} and the distributed training of graph embeddings\footnote{\url{https://github.com/LJeub/Local2Global_embedding}} are available from GitHub. This paper extends preliminary proof-of-concept results that were published as an extended abstract at the KDD 2021 ``Deep Learning on Graphs'' workshop \citep{jeub2021local2global}.

\section{Related work}
\label{sec:related}

This section briefly surveys scalable techniques for GNNs, and describes challenges arising when considering large scale graphs.  
Consider a graph $G(V, E)$ with $n = {\left\lvert V \right\rvert}$ nodes and $m=\abs{E}$ edges.
Let $\mat{A}$ be the adjacency matrix of $G$, where
\[
A_{ij} = \begin{cases} 1, & (i, j) \in E\\ 0, &\text{otherwise} \end{cases}
\]
then a general GNN layer is a differentiable function $\mat{H}=\operatorname{GNN}(\mat{X}; \mat{A})$, where $\mat{X} \in \R^{n\times d_\text{in}}$ are the input node features and $\mat{H} \in \R^{n \times d_\text{out}}$ are the output node features.
Importantly, $\operatorname{GNN}$ should be equivariant under node permutations, i.e., $\operatorname{GNN}(\mat{P}\mat{X}; \mat{P} \mat{A} \mat{P}^T) = \mat{P} \operatorname{GNN}(\mat{X}; \mat{A})$ for any $n \times n$ permutation matrix $\mat{P}$.
Let $\vec{x}_i$ be the $i$th row of $\mat{X}$ and similarly let $\vec{h}_i$ be the $i$th row of $\mat{H}$.
Most GNN layers can be written as
\begin{equation}
\vec{h}_i = \phi\left(\vec{x}_i, \bigoplus_{j=1}^n C(\mat{A})_{ij} \psi(\vec{x}_i, \vec{x}_j)\right), 
\label{eq:message_passing}
\end{equation}
where $\phi$ and $\psi$ are differentiable, vector-valued functions (typically with trainable parameters), $C(\mat{A})$ is the convolution matrix\footnote{Note that $C(\mat{A})$ should satisfy $C(\mat{P} \mat{A} \mat{P}^T) = \mat{P} C(\mat{A})\mat{P}^T$ for any $n\times n$ permutation matrix $\mat{P}$.}, and $\bigoplus$ is a permutation-invariant aggregator function (such as sum, max, \ldots).
The formulation in \cref{eq:message_passing} is often referred to as message passing where $\psi(\vec{x}_i, \vec{x}_j)$ is the message from node $j$ to node $i$.
Many different architectures have been proposed following this general framework (see~\cite{Wu2021} for a recent survey).

A particularly simple architecture which still has competitive performance \citep{Shchur2019} is GCN \citep{Kipf2017}.
Here, $C(\mat{A})$ is the normalised adjacency matrix with self-loops, i.e.,
\[
C(\mat{A}) = \bar{\mat{A}} =  \tilde{\mat{D}}^{-\frac{1}{2}}\tilde{\mat{A}}\tilde{\mat{D}}^{-\frac{1}{2}}, 
\]
where $\tilde{\mat{A}} = \mat{A}+\mat{I}$ and $\tilde{\mat{D}}$ is the diagonal matrix with entries $\tilde{\mat{D}}_{ii} = \sum_j \tilde{\mat{A}}_{ij}$.
Further, $\psi$ is a linear function, i.e., $\psi(x_j) = x_j^T \mat{W}$ where $\mat{W}\in \R^{d_\text{in} \times d_\text{out}}$ is a learnable parameter matrix.
The GCN layer can be written in matrix form as
\begin{equation}
H= \operatorname{GCN}(\mat{X}; \mat{A}) = \sigma\left( \bar{\mat{A}} \mat{X} \mat{W} \right),
\label{eq:GCN}
\end{equation}
where $\sigma$ is an element-wise non-linearity (typically ReLU). 

The key scalability problems for GNNs only concern deep architectures where we have $l$ nested GNN layers, i.e., 
\begin{equation}
H = \underbrace{\operatorname{GNN}(\operatorname{GNN}(\ldots; \mat{A}); \mat{A})}_l\label{eq:deep_GCN}.
\end{equation}

In particular, a single-layer GNN is easy to train in a scalable manner using mini-batch stochastic gradient descent (SGD). For simplicity, assume that we have a fixed feature dimension $d$, i.e., $d_\text{in}=d_\text{out}=d$ for all layers.
The original GCN paper of \cite{Kipf2017} uses full-batch gradient descent to train the model which entails the computation of the gradient for all nodes before updating the model parameters.
This is efficient in terms of time complexity per epoch ($O(lmd + lnd^2)$),  where $n$ is the number of nodes and $m$ is the number of edges. However, it requires storing all the intermediate embeddings and thus has memory complexity $O(lnd + ld^2)$. Further, as there is only a single parameter update per epoch, convergence tends to be slow.

The problem with applying vanilla mini-batch SGD (where we only compute the gradient for a sample of nodes, i.e., the batch) to a deep GNN model is that the embedding of the nodes in the final layer depends on the embedding of all the neighbours of the nodes in the previous layer and so on iteratively.
Therefore, the time complexity for a single mini-batch update approaches that for a full-batch update as the number of layers increases, unless the network has disconnected components.
There are mainly three families of methods \citep{Chen2020,Chiang2019}
that have been proposed to make mini-batch SGD training more efficient for GNNs.

\medskip  

\begin{description}[leftmargin=0pt, itemsep=\topsep]
\item[Layer-wise sampling.]
The idea behind layer-wise sampling is to sample a set of nodes for each layer of the nested GNN model and compute the embedding for sampled nodes in a given layer only based on embeddings of sampled nodes in the previous layer, rather than considering all the neighbours as would be the case for vanilla SGD. This seems to have first been used by GraphSAGE \citep{Hamilton2018}, where a fixed number of neighbours is sampled for each node at each layer.
However, this results in a computational complexity that is exponential in the number of layers and also redundant computations as the same intermediate nodes may be sampled starting from different nodes in the batch.
Later methods avoid the exponential complexity by first sampling a fixed number of nodes for each layer either independently (FastGCN, \cite{Chen2018a}) or conditional on being connected to sampled nodes in the previous layer (LADIES, \cite{Zou2019}) and reusing embeddings.
Both methods use importance sampling to correct for bias introduced by non-uniform node-sampling distributions.
Also notable is \cite{Chen2018b}, which uses variance reduction techniques to effectively train a GNN model using neighbourhood sampling as in GraphSAGE with only 2 neighbours per node.
However, this is achieved by storing hidden embeddings for all nodes in all layers and thus has the same memory complexity as full-batch training.

\medskip  

\item[Linear model.]
Linear models remove the non-linearities between the different GNN layers which means that the model can be expressed as a single-layer GNN with a more complicated convolution operator and hence trained efficiently using mini-batch SGD.
Common choices for the convolution operator are powers of the normalised adjacency matrix \citep{Wu2019a} and variants of personalised Page-Rank (PPR) matrices \citep{Busch2020,Chen2020,Bojchevski2020,Klicpera2019}.
Another variant of this approach is \cite{frasca2020sign}, which proposes combining different convolution operators in a wide rather than deep architecture.
There are different variants of the linear model architecture, depending on whether the non-linear feature transformation is applied before or after the propagation (see \cite{Busch2020} for a discussion), leading to predict-propagate and propagate-predict architectures respectively.
The advantage of the propagate-predict architecture is that one can pre-compute the propagated node features (e.g., using an efficient push-based algorithm as in \cite{Chen2020}) which can make training highly scalable.
The disadvantage is that this will densify sparse features which can make training harder \citep{Bojchevski2020}.
However, the results from \cite{Busch2020} suggest that there is usually not much difference in prediction performance between these options (or the combined architecture where trainable transformations are applied before and after propagation).

\medskip  

\item[Subgraph sampling.]
Subgraph sampling techniques \citep{Zeng2019,Chiang2019,Zeng2020} construct batches by sampling an induced subgraph of the full graph.
In particular, for subgraph sampling methods, the sampled nodes in each layer of the model in a batch are the same.
In practice, subgraph sampling seems to outperform layer-wise sampling \citep{Chen2020}.
GraphSAINT \citep{Zeng2020}, which uses a random-walk sampler with an importance sampling correction similar to \cite{Chen2018a,Zou2019}, seems to have the best performance so far.
Our {\ltg} approach shares similarities with subgraph sampling, most notably ClusterGCN \citep{Chiang2019}, which uses graph clustering techniques to sample the batches.
The key distinguishing feature of our approach is that we train independent models for each patch (and then we stitch together the resulting embeddings),  whereas for ClusterGCN, model parameters have to be kept synchronised 
for different batches, which hinders fully distributed training and its associated key benefits (see~\cref{sec:introduction}). Another recent method, \citep[GNNAutoScale,][]{Fey2021}, combines subgraph sampling with historical embeddings for the out-of-sample neighbors of nodes in the sampled subgraph. Historical embeddings avoid the loss of information inherent in subgraph sampling techniques, but require storing all intermediate embeddings for all nodes, resulting in a large memory footprint. While one can use slower memory for the historical embeddings without significantly impacting training speed, keeping historical embeddings synchronised will still result in significant communications overhead in a distributed setting.
\end{description}

\section{The  {\ltg} algorithm}\label{sec:l2g}

The key idea behind the {\ltg} approach to graph embedding is to embed different parts of a graph independently by splitting the graph into overlapping ``patches'' and then stitching the patch node embeddings together to obtain a single global node embedding for each node. 
The stitching of the patch node embeddings proceeds by estimating the orthogonal transformations (i.e., rotations and reflections) and translations for the embedding patches that best aligns them based on the overlapping nodes.

Consider a graph $G(V, E)$ with node set $V$ and edge set $E$.
The input for the {\ltg} algorithm is a patch graph $G_p(\set{P}, E_p)$ together with a set of patch node embeddings $\set{X} = \{\mat{X}^{(k)}\}_{k=1}^p$. The node set of the patch graph is the set of patches $\set{P}$, where each node  $P_k \in \set{P}$ of the patch graph (i.e., a ``patch'') is a subset of $V$ and has an associated node embedding $\mat{X}^{(k)} \in \R^{\abs{P_k}\times d}$. The edge set $E_p$ of the patch graph identifies the pairs of patches for which we have an estimate of the relative transformation.

We require that the set of patches $\set{P}=\{P_k\}_{k=1}^p$ is a cover of the node set $V$ (i.e., $\bigcup_{k=1}^p P_k = V$), and that the patch embeddings all have the same dimension $d$.
We further assume that the patch graph is connected and that
the patch edges satisfy the minimum overlap condition\footnote{Note that a pair of patches $P_i, P_j$ that satisfies the minimum overlap condition $\abs{P_i \cap P_j} \geq d+1$ is not necessarily connected in the patch graph, i.e., we do not necessarily estimate the relative transformations for all pairs of patches that satisfy the minimum overlap criterion.} $\{P_i, P_j\} \in E_p \implies \abs{P_i \cap P_j} \geq d+1$. This condition ensures that we can estimate the relative orthogonal transformation for a pair of connected patches.

\begin{figure}
\centering
\includegraphics[width=0.8\linewidth, trim={1.05cm 0 0 1.8cm}, clip]{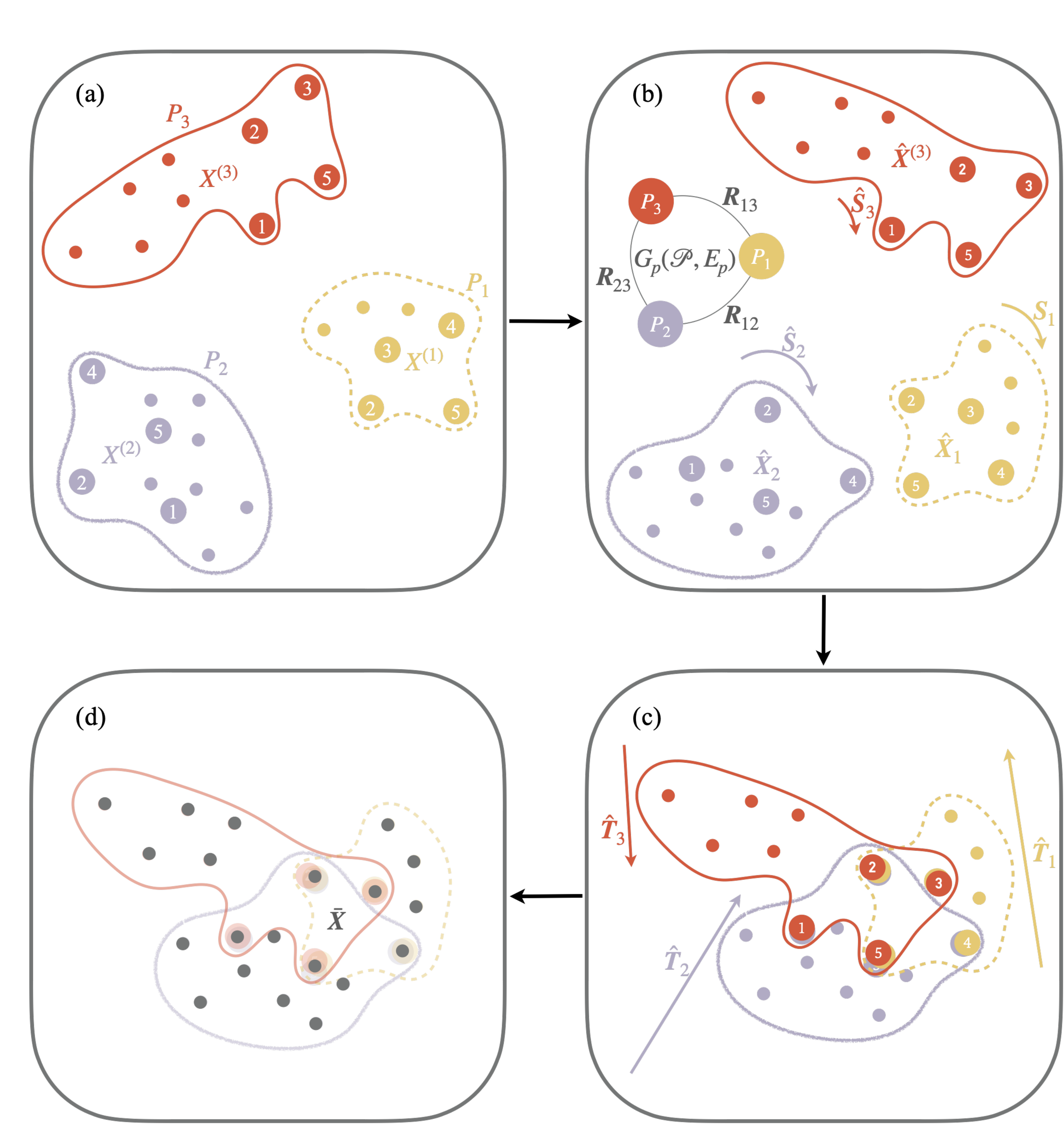}

\caption{Schematic overview of the different steps of the local2global algorithm. (a) Independent patch embeddings with identified overlapping nodes. (b) Synchronisation over orthogonal transformations. (c) Synchronisation over translations. (d) Global node embedding as centroid of aligned patch embeddings.}
\label{fig:l2g_schematic}
\end{figure}

The {\ltg} algorithm for aligning the patch embeddings proceeds in three stages and is an evolution of the approach in \cite{Cucuringu2012a,Cucuringu2012b}.
We assume that each patch embedding $\mat{X}^{(k)}$ is a perturbed version of an underlying global node embedding $\mat{X}$, where the perturbation is composed of scaling ($\R_+$), reflection ($Z_2$), rotation (SO($d$)), translation ($\mathbb{R}^d$), and noise.
The goal is to estimate the transformation applied to each patch using only pairwise noisy measurements of the relative transformation for pairs of connected patches.
In the first two stages, we estimate the scale and the orthogonal transformation to apply to each patch embedding, using a variant of the eigenvector synchronization method \citep{Singer2011,Cucuringu2012a,Cucuringu2012b}.
In the second stage, we estimate the patch translations by solving a least-squares problem. 
Note that unlike \cite{Cucuringu2012a,Cucuringu2012b}, we solve for translations at the patch level rather than solving a least-squares problem for the node coordinates.
This means that the computational cost for computing the patch alignment is independent of the size of the original network and depends only on the amount of patch overlap, the number of patches and the embedding dimension.
We illustrate the different steps of the {\ltg} algorithm in \cref{fig:l2g_schematic}.

\subsection{Eigenvector synchronisation over scales}

As an optional first step of the {\ltg} algorithm, we synchronise the scales of the patch embeddings. Whether or not synchronisation over scales is beneficial depends on the properties of the chosen embedding method. In particular, if the embedding method determines the overall scale based on global properties of the graph (e.g., through normalisation), scale synchronisation is likely beneficial as one expects patch embeddings to have different scales. However, if the embedding method determines the overall scale only based on local properties (e.g., by trying to fix edge lengths), scale synchronisation may not be necessary and may hurt the final embedding quality by fitting noise.
We assume that to each patch $P_i$, there corresponds an unknown scale factor $s_i \in \R_+$. For each pair of connected patches $(P_i, P_j) \in E_p$, we have a noisy estimate $r_{ij} \approx s_i s_j^{-1}$ for the relative scale of the two patches. We estimate $r_{ij}$ using the method from \cite{Horn1988} as 
\begin{equation}
    r_{ij} = \frac{\sqrt{\sum_{u \in P_i \cap P_j} X^{(i)}_u - \frac{1}{\abs{P_i \cap P_j}}\sum_{v \in P_i \cap P_j} X^{(i)}_v}}{\sqrt{\sum_{u \in P_i \cap P_j} X^{(j)}_u - \frac{1}{\abs{P_i \cap P_j}}\sum_{v \in P_i \cap P_j} X^{(j)}_v}}\,.
\end{equation}

For each pair of connected patches $(P_i, P_j) \in E_p$, we have a consistency equation $s_i \approx r_{ij} s_j$, as we can estimate the scale of a patch based on the scale of one of its neighbors in the patch graph and the relative scale between them. Using a weighted average to combine estimates based on different neighbors, we arrive at 
\begin{equation}
    s_i \approx \sum_j \frac{w_{ij} r_{ij}s_j}{\sum_j w_{ij}}, \label{eq:scale_consistency}
\end{equation}
where we set $w_{ij} = \abs{P_i \cap P_j}$ if $\{P_i, P_j\} \in E_p$ and $w_{ij}=0$ otherwise, as we expect pairs of patches with larger overlap to result in a more reliable estimate of the relative scale. Based on \cref{eq:scale_consistency}, we use the leading eigenvector $\hat{\vec{s}}$ of the matrix with entries 
\[
\frac{w_{ij} r_{ij}}{\sum_j w_{ij}}, 
\]
(which has strictly positive entries by the Perron–Frobenius theorem, provided the patch graph is connected) to estimate the scale factors. We normalise $\hat{\vec{s}}$ such that $\frac{1}{p} \sum_j \hat{s}_j=1$ to fix the global scale and compute the scale-synchronized patch embeddings as $\tilde{\mat{X}}^{(i)} = \mat{X}^{(i)}\hat{s}_i$.

\subsection{Eigenvector synchronisation over orthogonal transformations} 

Synchronisation over orthogonal transformations proceeds in a similar way to synchronisation over scales. 
We assume that to each patch $P_i$, there corresponds an unknown group element $S_i \in O(d) \simeq Z_2 \times SO(d)$ (represented by a $d \times d$ orthogonal matrix), and for each pair of connected patches $(P_i, P_j) \in E_p$ we have a noisy proxy for $S_i S_j^{-1}$, which is precisely the setup of the \textit{group synchronization} problem.

For a pair of connected patches $P_i, P_j \in \set{P}$ such that $\{P_i, P_j\} \in E_p$ we can estimate the relative  rotation/reflection via a Procrustes alignment, by applying the method from \cite{Horn1988}\footnote{Note that the rotation/reflection can be estimated without knowing the relative translation.} to their overlap as $\abs{P_i \cap P_j} \geq d+1$.
Thus, we can construct a block matrix $\mat{R}$ where $\mat{R}_{ij}$ is the $d \times d$ orthogonal matrix representing the estimated relative transformation from patch $P_j$ to patch $P_i$ if $\{P_i, P_j\} \in E_p$ and $\mat{R}_{ij}=\mat{0}$ otherwise, such that $\mat{R}_{ij} \approx \mat{S}_i \mat{S}_j^{T}$ for connected patches.

In the noise-free case, we have the consistency equations 
$\mat{S}_i = \mat{R}_{ij} \mat{S}_j$
for all $i, j$ such that $\{P_i, P_j\} \in E_p$.
One can combine the consistency equations for all neighbours of a patch to arrive at 
\begin{equation}
\mat{S}_i = \sum_j \mat{M}_{ij}\mat{S}_j, \qquad \mat{M}_{ij} = \frac{w_{ij} \mat{R}_{ij}}{\sum_{j} w_{ij}},
\label{eq:syn_sum}
\end{equation}
where we use $w_{ij} = \abs{P_i \cap P_j}$ to weight the contributions, as we expect a larger overlap to give a more robust estimate of the relative transformation. 
We can write \cref{eq:syn_sum} as 
$\mat{S} = \mat{M} \mat{S}$, where $\mat{S} = (\mat{S}_1, \ldots, \mat{S}_p)^T$ is a $pd \times d$ block-matrix and $\mat{M}$ is a $pd \times pd$ block-matrix. 
Thus, in the noise-free case, the columns of $\mat{S}$ are eigenvectors of $\mat{M}$ with eigenvalue 1.
Thus, following \cite{Cucuringu2012a,Cucuringu2012b}, we can use the $d$ leading eigenvectors\footnote{While $\mat{M}$ is not symmetric, it is similar to a symmetric matrix and thus has real eigenvectors.} of $\mat{M}$ as the basis for estimating the transformations.
Let $\mat{U}= (\mat{U}_1, \ldots, \mat{U}_p)^T$ be the $pd \times d$ matrix whose columns are the $d$ leading eigenvectors of $\mat{M}$, where $\mat{U}_i$ is the $d \times d$ block of $\mat{U}$ corresponding to patch $P_i$.
We obtain the estimate $\hat{\mat{S}}_i$ of $\mat{S}_i$ by finding the nearest orthogonal transformation to $\mat{U}_i$ using an SVD \citep{Horn1988}, and hence the estimated rotation-synchronised embedding of patch $P_i$ is $\hat{\mat{X}}^{(i)} = \tilde{\mat{X}}^{(i)} \hat{\mat{S}}_i^T$.

\subsection{Synchronisation over translations}

After synchronising the rotation of the patches, we can estimate the translations by solving a least-squares problem.
Let $\hat{\mat{X}}_i^{(k)} \in \R^d$ be the (rotation-synchronised) embedding of node $i$ in patch $P_k$ ($\hat{\mat{X}}_i^{(k)}$ is only defined if $i\in P_k$).

Let $\mat{T}_k\in \R^d$ be the translation of patch $k$; then, in the noise-free case, the following consistency equations hold true
\begin{equation}
\hat{\mat{X}}_i^{(k)} + \mat{T}_k = \hat{\mat{X}}_i^{(l)} + \mat{T}_l,\qquad i \in P_k \cap P_l .\label{eq:consistency_single}
\end{equation}
We can combine the conditions in \cref{eq:consistency_single} for each edge in the patch graph to arrive at 
\begin{equation}
\mat{B} \mat{T} = \mat{C}, \qquad \mat{C}_{(P_k, P_l)} = \frac{\sum_{i \in P_k \cap P_l} \hat{\mat{X}}_i^{(k)} - \hat{\mat{X}}_i^{(l)}}{\abs{P_k \cap P_l}},  \label{eq:consistency_combined}
\end{equation}
where $\mat{T} \in \R^{\abs{\set{P}} \times d}$ is the matrix such that the $k$th row of $\mat{T}$ is the translation $\mat{T}_k$ of patch $P_k$ and $\mat{B}\in \{-1,1\}^{\abs{E_p}\times \abs{\set{P}}}$ is the incidence matrix of the patch graph with entries $\mat{B}_{(P_k,P_l),j} = \delta_{lj}-\delta_{kj}$, where $\delta_{ij}$ denotes the Kronecker delta. 
\Cref{eq:consistency_combined} defines an overdetermined linear system that has the true patch translations as a solution in the noise-free case. 
In the practical case of noisy patch embeddings, we can instead solve \cref{eq:consistency_combined} in the least-squares sense
\begin{equation}
\hat{\mat{T}} = \argmin_{\mat{T} \in \R^{p\times d}} \norm{\mat{B}\mat{T} - \mat{C}}_2^2.
\label{eq:translation}
\end{equation}
We estimate the aligned node embedding $\bar{\mat{X}}$ in a final step using the centroid of the aligned patch embeddings of a node, i.e.,
\begin{equation}
\bar{\mat{X}}_i = \frac{\sum_{\{P_k \in \set{P} \given i \in P_{k}\}} \hat{\mat{X}}_i^{(k)}+\hat{\mat{T}}_k}{\abs{\{P_k \in \set{P} \given i \in P_{k}\}}}. \label{eq:mean_embedding}
\end{equation}

\subsection{Scalability of the {\ltg} algorithm}

The patch alignment step of {\ltg} is highly scalable and does not directly depend on the size of the input data.
The cost for computing the matrix $M$ is $O(\abs{E_p}od^2)$ where $o$ is the average overlap between connected patches (typically $o \sim d$) and the cost for computing the vector $b$ is $O(\abs{E_p}od)$.
Both operations are trivially parallelisable over patch edges.
The translation problem can be solved with an iterative least-squares solver with a per-iteration complexity of $O(\abs{E_p}d)$.
The limiting step for {\ltg} is usually the synchronization over orthogonal transformations which requires finding $d$ eigenvectors of a $d\abs{\set{P}} \times d\abs{\set{P}}$ sparse matrix with $\abs{E_p}d^2$ non-zero entries for a per-iteration complexity of $O(\abs{E_p}d^3)$.
This means that in the typical scenario where we want to keep the patch size constant, the patch alignment scales almost linearly with the number of nodes in the dataset, as we can ensure that the patch graph remains sparse, such that $\abs{E_p}$ scales almost linearly with the number of patches.
The $O(\abs{E_p}d^3)$ scaling puts some limitations on the embedding dimension attainable with the {\ltg} approach, though, as we can see from the experiments  in \cref{subsec:embeddings}, it remains feasible for reasonably high embedding dimension. We note that one can use a hierarchical version of {\ltg} (see \cref{sec:hierchical}) to further increase the embedding dimension in a scalable fashion.


The preprocessing to divide the network into patches scales as $O(m)$.
The speed-up attainable due to training patches in parallel depends on the oversampling ratio (i.e., the total number of edges in all patches divided by the number of edges in the original graph).
As seen in \cref{subsec:embeddings}, we achieve good results with moderate oversampling ratios.

\subsection{Hierarchical {\ltg} algorithm}
\label{sec:hierchical}

For very large data sets, or in other settings where we have a large number of patches, solving the eigenvalue problem implied by \cref{eq:syn_sum} to synchronise the orthogonal transformations becomes a computational bottleneck. We implement a hierarchical approach for the patch alignment step to address  this issue. In the hierarchical approach, we cluster the patch graph and apply {\ltg} to align the patches within each cluster and compute the aligned node embedding for each cluster. We then apply {\ltg} again to align the patch clusters embeddings. Such a hierarchical approach would make it trivial to parallelize the patch alignment,  but may come with a penalty in terms of the embedding quality. 

\section{Evaluation on edge reconstruction and semi-supervised classification}
\label{sec:semi-supervised}
\subsection{Data sets}

To test the viability of the {\ltg} approach to graph embeddings, we consider data sets of increasing node set size.
For the small-scale tests, we consider the Cora and Pubmed citation data sets from~\cite{Yang2016} and the Amazon photo co-purchasing data set from \cite{Shchur2019}.
We consider only nodes and edges in the largest connected component (LCC) for the small-scale tests.
For the large-scale test we use the MAG240m data from \cite{hu2021ogblsc}.
We consider only the citation edges and preprocess the data set to make the edges undirected.
We use the features as is without further processing.
We show some statistics of the data sets in \cref{tab:data}.

\begin{table}[h]
\caption{Data sets\label{tab:data}}
\sisetup{group-minimum-digits=4}
\begin{tabular}{r|S[table-number-alignment=right,table-format=9.0]S[table-number-alignment = right,table-format=10.0]S[table-number-alignment = right,table-format=4.0]S[table-number-alignment = right,table-format=3.0]}
& {nodes} & {edges} & {features} & {classes}\\
\hline
Cora \citep{Yang2016} & 2485 & 10138 & 1433 & 7\\
Amazon photo \citep{Shchur2019} & 7487 & 238086 & 745 & 8\\
Pubmed \citep{Yang2016} & 19717 & 88648 & 500 & 3 \\
MAG240m \citep{hu2021ogblsc} & 121751666 & 2593241212 & 768 & 153
\end{tabular}

\end{table}

\subsection{Patch graph construction}

\begin{algorithm2e}[tb]
\caption{Sparsify patch graph\label{alg:sparsify}}
\KwIn{$G_p(\set{P}, E_p)$, $G(V, E)$, target patch degree $k$}
\KwResult{sparsified patch graph $G_p(\set{P}, \tilde{E}_p)$}
\ForEach{$\{P_i, P_j\} \in E_p$}{
Compute conductance weight
\begin{algomathdisplay}
c_{ij} = \tfrac{\abs{\{(u, v) \in E \given u \in P_i, v \in P_j\}}}{\min(\abs{\{(u, v) \in E \given u \in P_i\}}, \abs{\{(u, v) \in E \given u \in P_j\}})}
\end{algomathdisplay}
}
\ForEach{$\{P_i, P_j\} \in E_p$}{
Compute effective resistance $r_{ij}$ between $P_i$ and $P_j$ in $G_p(\set{P}, E_p, c)$ using the algorithm of \cite{Spielman2011a}\;
Let $w_{ij} = r_{ij} c_{ij}$\;
}
Initialize $\tilde{E}_p$ with a maximum spanning tree of $G_p(\set{P}, E_p, w)$\;
Sample the remaining $(k-1)p + 1$ edges from $E_p \setminus \tilde{E}_p$ without replacement and add them to $\tilde{E}_p$, where edge $\{P_i, P_j\}$ is sampled with probability $w_{ij}$\;
\Return $G_p(\set{P}, \tilde{E}_p)$
\end{algorithm2e}

\begin{algorithm2e}[tb]
\caption{Create overlapping patches\label{alg:overlap}}
\KwIn{$\set{C}$, $E_p$, $G(V, E)$, min overlap $l$, max overlap $u$}
\KwResult{Overlapping patches $\set{P}$}
Initialise $\set{P} = \set{C}$\;
Define the neighbourhood of a set of nodes $U$ as
\begin{algomathdisplay}
N(U) = \{j \given i \in U, (i, j) \in E, j \notin U\}
\end{algomathdisplay}
\ForEach{$P_i \in \set{P}$}{
\ForEach{$P_j$ s.t. $\{P_i, P_j\} \in E_p$}{
	Let $F = N(C_i) \cap C_j$\;
	\While{$\abs{P_i \cap C_j} < l/2$}{
		\If{$\abs{F} + \abs{P_i \cap C_j} > u/2$}{
			reduce $F$ by sampling uniformly at random such that $\abs{F} = u/2 - \abs{P_i \cap C_j}$\;}
		Let $P_i = P_i \cup F$\;
		Let $F = (N(F) \cap C_j) \setminus P_i$\;
		}
	}
}
\Return{$\set{P}$}
\end{algorithm2e}

The first step in the {\ltg} embedding pipeline is to divide the network $G(V, E)$ into overlapping patches. 
In some federated-learning applications, the network may already be partitioned and some or all of the following steps may be skipped provided the resulting patch graph is connected and satisfies the minimum overlap condition for the desired embedding dimension. 
Otherwise, we proceed by first partitioning the network into non-overlapping clusters and then enlarging clusters to create overlapping patches. 
This two-step process makes it easier to ensure that patch overlaps satisfy the conditions for the {\ltg} algorithm without introducing excessive overlaps,  than if we were to use a clustering algorithm that produces overlapping clusters directly. We use the following pipeline to create the patches:
\begin{itemize}
\item \textbf{Partition the network} into $p$ non-overlapping clusters $\set{C}=\{C_k\}_{k=1}^p$ such that $\abs{C_k} \geq \frac{d+1}{2}$ for all $k$.
We use METIS \citep{Karypis1998} to cluster the networks for the small-scale  experiments in \cref{subsec:embeddings}.
For the large-scale experiments using MAG240m, we use FENNEL \citep{Tsourakakis2014} to cluster the network.
FENNEL obtains high-quality clusters in a single pass over the nodes and is thus scalable to very large networks.

\item \textbf{Initialize the patches} to $\set{P} = \set{C}$ and define the patch graph $G_p(\set{P}, E_p)$, where $\{P_i, P_j\} \in E_p$ iff there exist nodes $i \in P_i$ and $j \in P_j$ such that $\{i, j\} \in E$. (Note that if $G$ is connected, $G_p$ is also connected.)

\item \textbf{Sparsify the patch graph} $G_p$ to have mean degree $k$ using \cref{alg:sparsify} adapted from the effective-resistance sampling algorithm of \cite{Spielman2011a}. For MAG240m, we sample edges based on the raw conductance values rather than the effective resistances as computing effective resistances for all patch edges is no longer computationally feasible at this scale. 

\item \textbf{Expand the patches} to create the desired patch overlaps.
We define a lower bound $l \geq d+1$ and upper bound $u$ for the desired patch overlaps and use \cref{alg:overlap} to expand the patches such that $\abs{P_i \cap P_j} \geq l$ for all $\{P_i, P_j\} \in E_p$.
\end{itemize}

\subsection{Embedding models}
We consider two embedding approaches with different objective functions, the variational graph auto-encoder (VGAE) architecture of \cite{Kipf2016} and deep-graph infomax (DGI) architecture of \cite{Velickovic2019}.  We reiterate however that one can a-priori incorporate any embedding method within  our pipeline (see \cref{sec:l2g}). 
We use the Adam optimizer \citep{Kingma2015} for training with learning rate set to 0.001 to train all models. We use early-stopping on the training loss with a miximum training time of \num{10000} epochs and a patience of 20 epochs as in \cite{Velickovic2019}\footnote{We use the same training strategy also for the VGAE models as we find that it gives a small improvement over the training strategy of \cite{Kipf2016}.}. For DGI, we normalize the bag-of-word features such that the features for each node sum to 1 as in \cite{Velickovic2019}. However, we do not apply the feature normalization when training VGAE models as it hurts performance.
We set the hidden dimension of the VGAE models to $4\times d$, where $d$ is the embedding dimension.

For the small-scale test, we repeat the training procedure 10 times with different random initialisations and keep the embedding with the best reconstruction AUC. For the large-scale test, we only use a single training run. 

\subsection{Evaluation tasks}

We consider two tasks for evaluating the  embedding quality. The first task we consider is network reconstruction, where one tries to reconstruct the edges based on the embedding. We use the inner product between embedding vectors as the scoring function and use the area under the ROC curve (AUC) as the quality measure. 

The second task we consider is semi-supervised classification. We follow the procedure of \cite{Velickovic2019} and train a logistic regression model to predict the class labels based on the embedding. For the small-scale tests we use 20 labelled examples per class as the training set, and report the average and standard deviation over 50 random train-test splits. For the large-scale test on MAG240m, we use the full training data to train the classifier and evaluate the performance on the validation set\footnote{The labels for the test set are not currently included in the public data.}, except for \num{10000} randomly sampled papers which are used as the validation set. In all cases, we train the classifier for a maximum of \num{10000} epochs with early stopping based on the validation loss with a patience of \num{20} epochs. We use the ADAM optimiser \citep{Kingma2015} with a learning rate of \num{0.01} for training.

\subsection{Results}
\label{subsec:embeddings}

\begin{figure}
	\subfloat[\label{fig:umap-cora}]{%
	\includegraphics[width=0.49\linewidth]{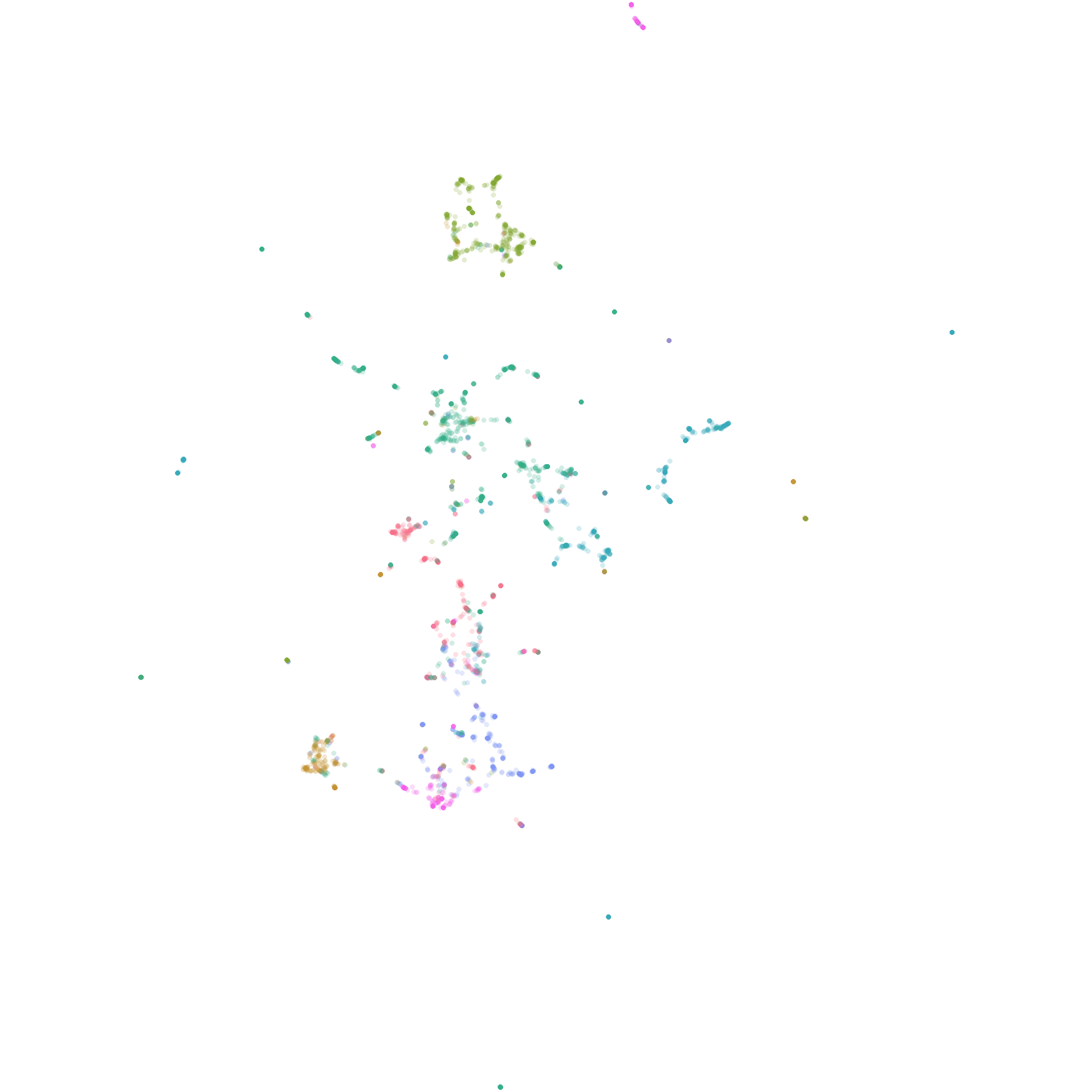}}\hfill
	\subfloat[\label{fig:umap-AMZ}]{%
	\includegraphics[width=0.49\linewidth]{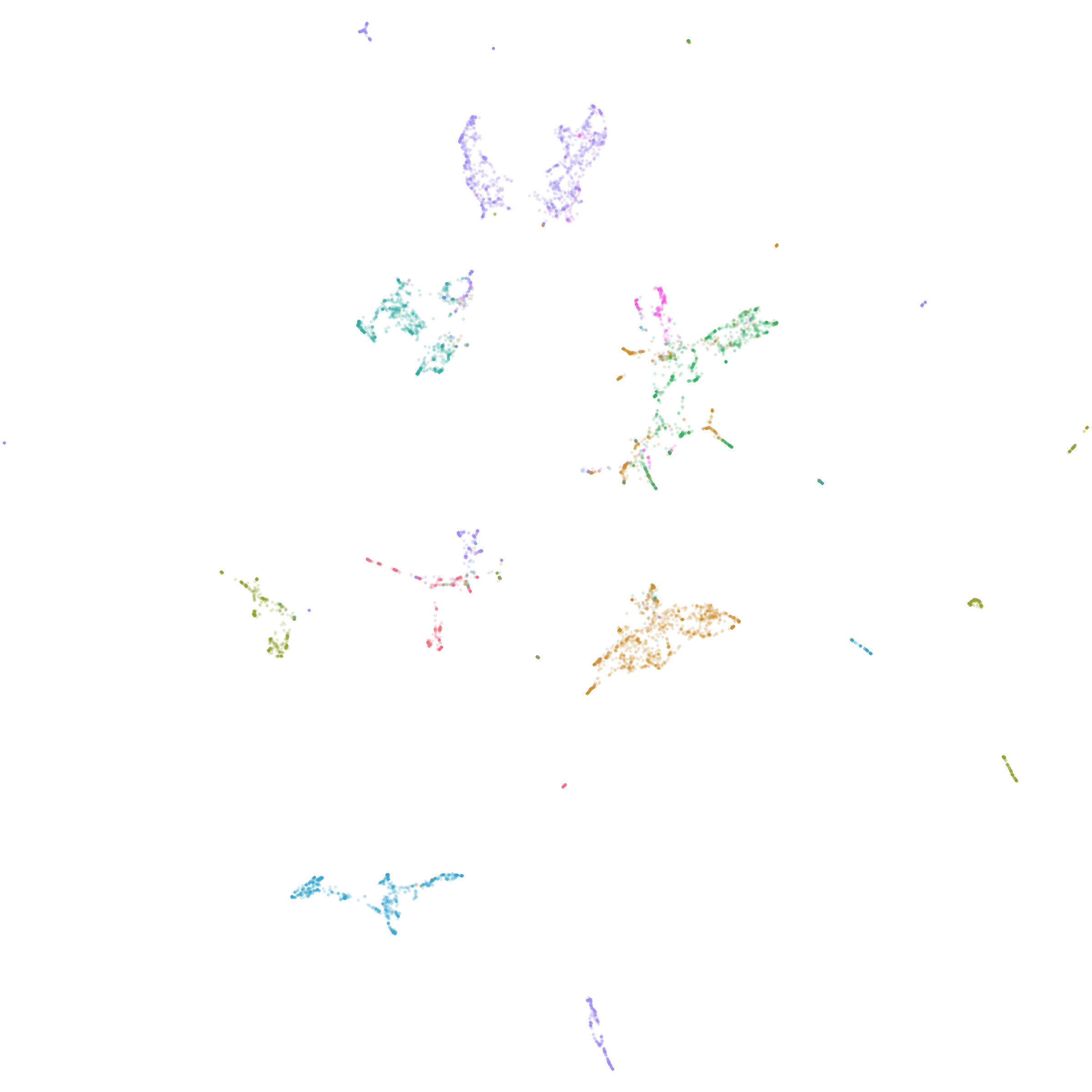}}\\

	\subfloat[\label{fig:umap-pubmed}]{%
	\includegraphics[width=0.49\linewidth]{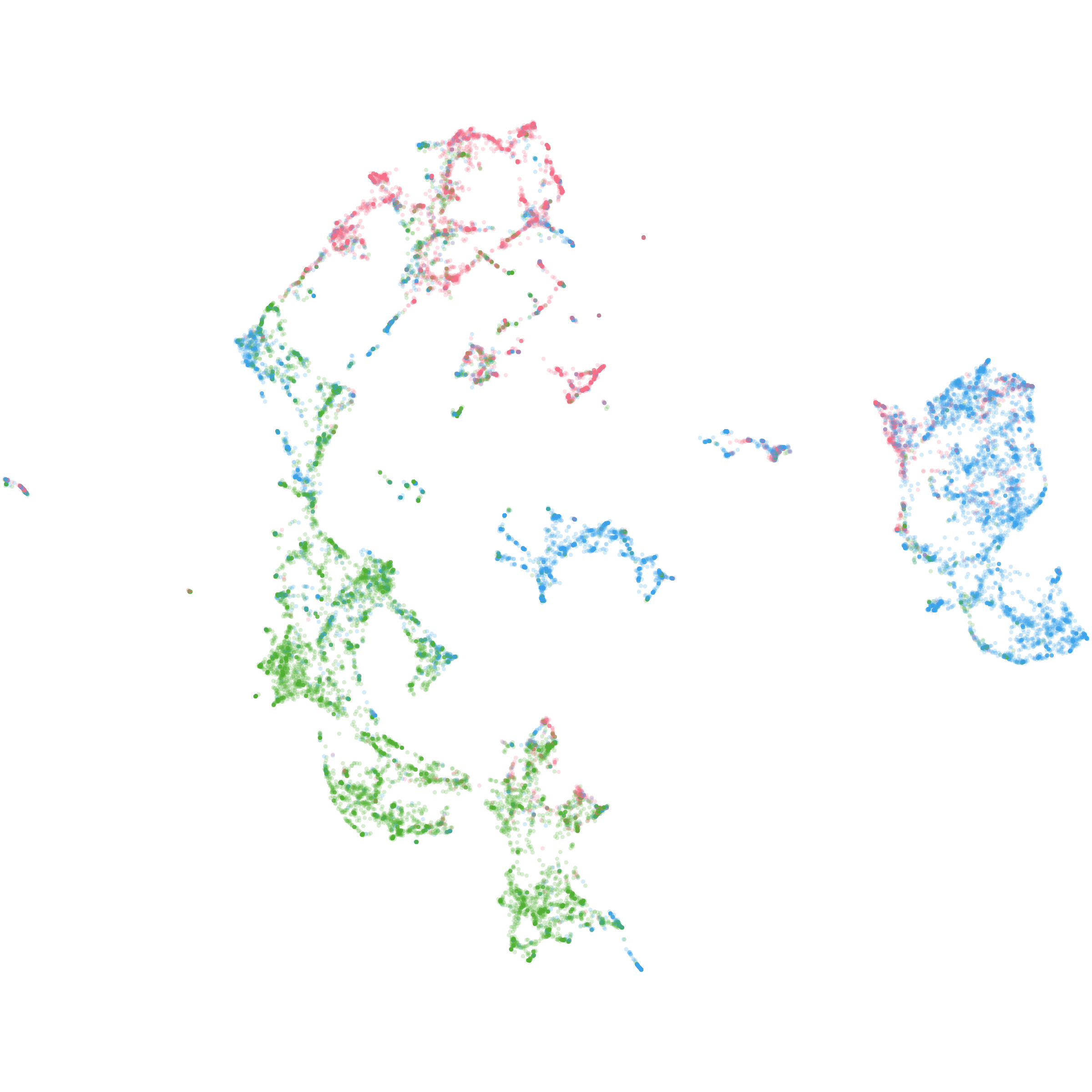}}\hfill
	\subfloat[\label{fig:umap-MAG240m}]{%
	\includegraphics[width=0.49\linewidth]{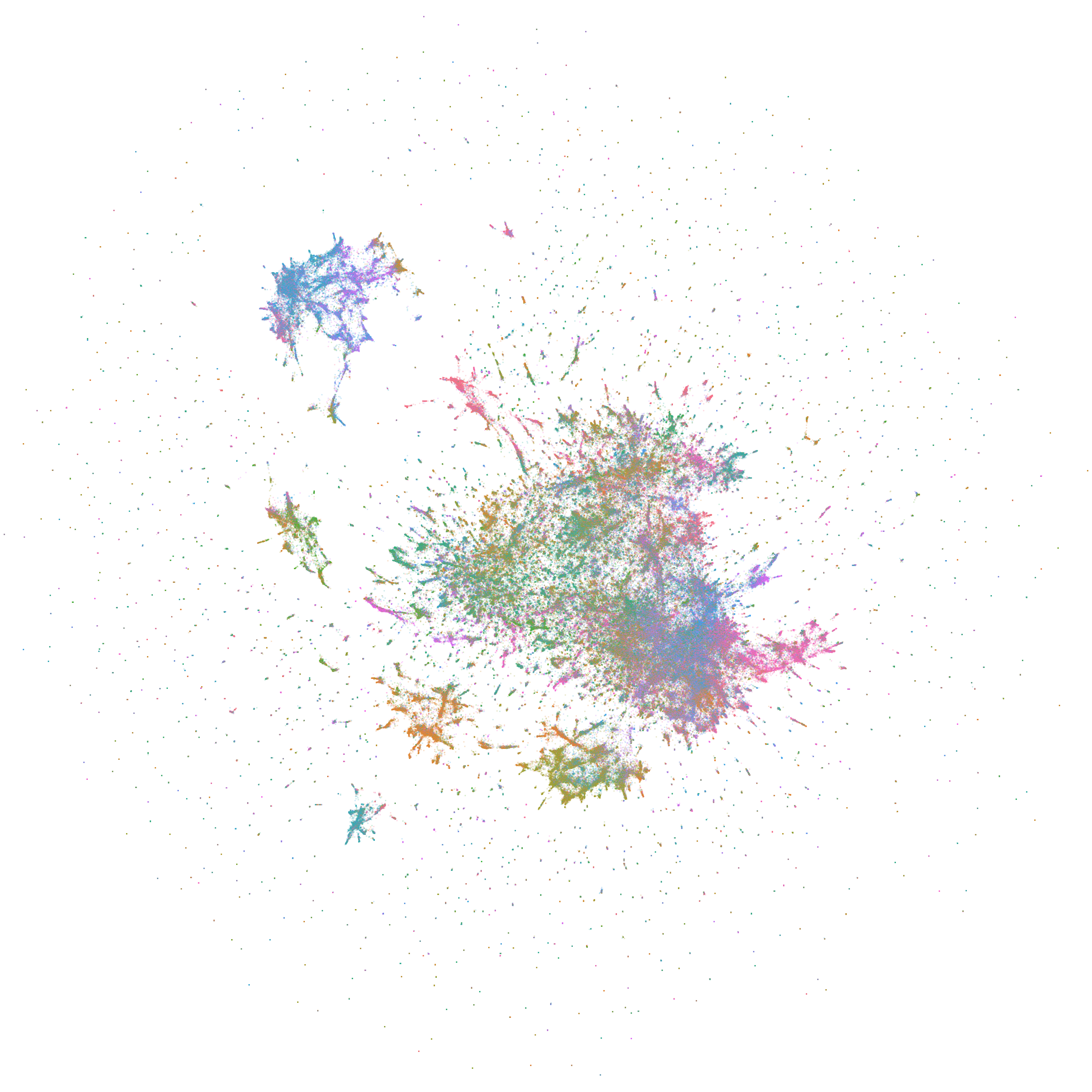}}
	\caption{UMAP projection of 128-dimensional VGAE-l2g embeddings for \protect\subref{fig:umap-cora} Cora, \protect\subref{fig:umap-AMZ} Amazon photo, \protect\subref{fig:umap-pubmed} Pubmed, and \protect\subref{fig:umap-MAG240m} MAG240m. Nodes are coloured by class label. For MAG240m, the visualisation is based on a sample of \num{500000} labeled  papers.\label{fig:umap}}
\end{figure}

To get some intuition about the embeddings we obtain, we first visualise the results using UMAP \citep{mcinnes2020umap}. We show the results for the $d=128$ stitched {\ltg} embeddings obtained with VGAE in \cref{fig:umap}. For the small-scale data sets, we observe that the different classes appear clustered in the embedding. The results for the large-scale test do not currently highlight any clearly visible underlying structure.  This may be partly due to the much larger number of classes (153), and may also suggest that the quality of the embedding is not as good.

\subsubsection{Small-scale tests}

\begin{figure*}
\subfloat[\label{fig:VGAE_auc_scores:cora}]{%
\begin{minipage}{0.33\linewidth}
\includegraphics[width=\linewidth, clip, trim={0 0.2cm 0 0.8cm}]{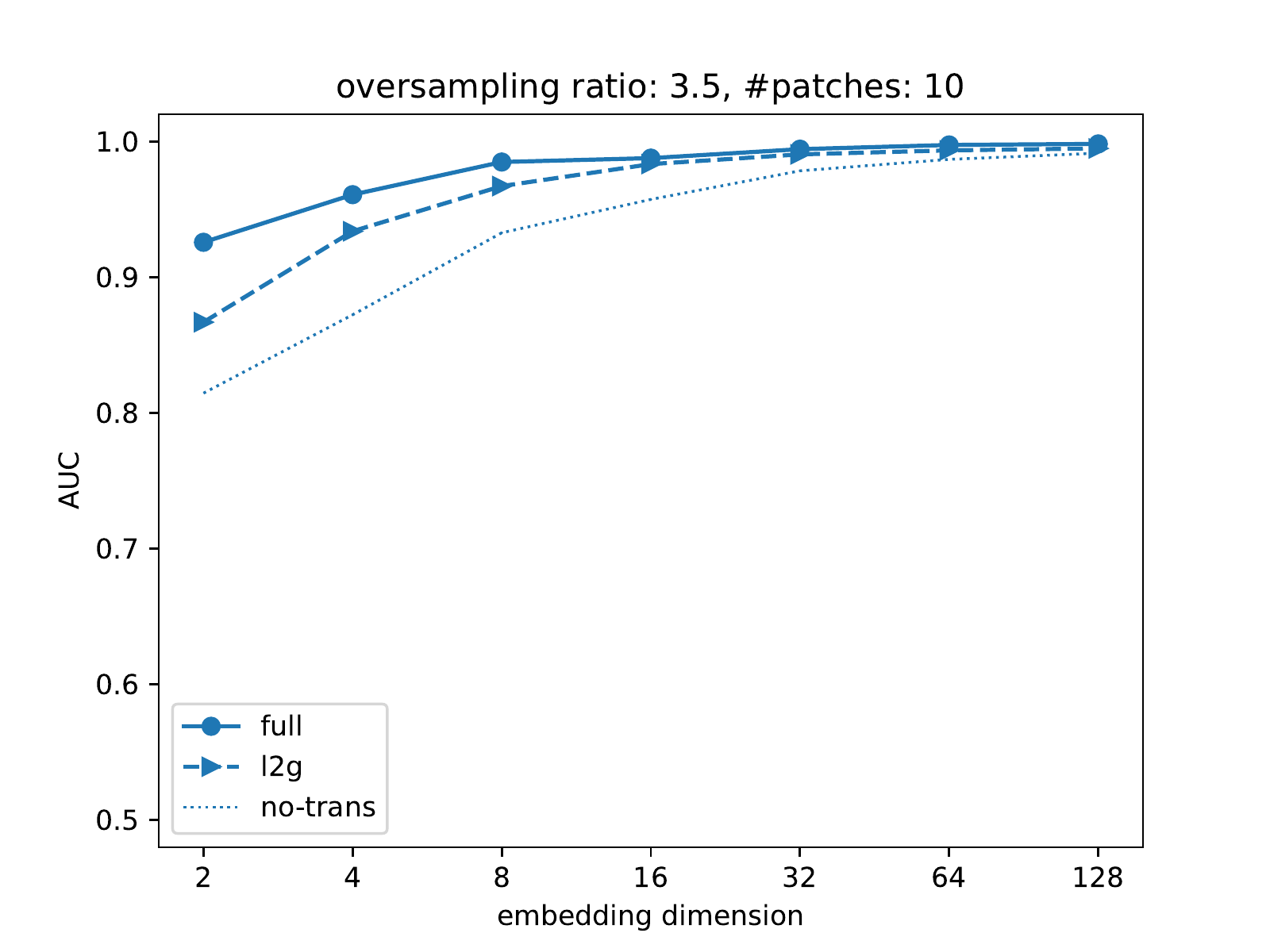}
\end{minipage}}\hfill
\subfloat[\label{fig:VGAE_auc_scores:amz_photo}]{%
\begin{minipage}{0.33\linewidth}
\includegraphics[width=\linewidth, clip, trim={0 0.2cm 0 0.8cm}]{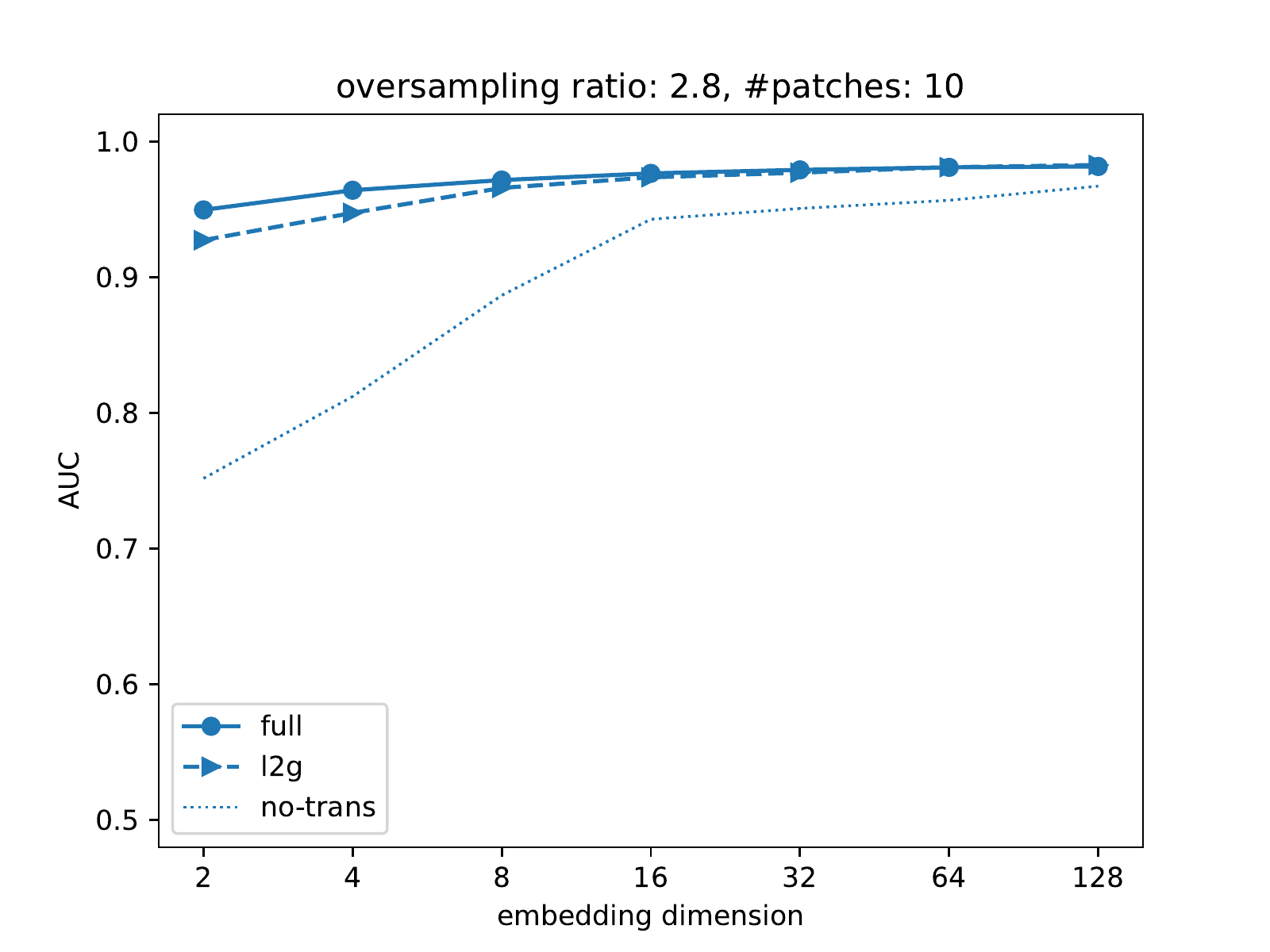}%
\end{minipage}}\hfill
\subfloat[\label{fig:VGAE_auc_scores:pubmed}]{%
\begin{minipage}{0.33\linewidth}
\includegraphics[width=\linewidth, clip, trim={0 0.2cm 0 0.8cm}]{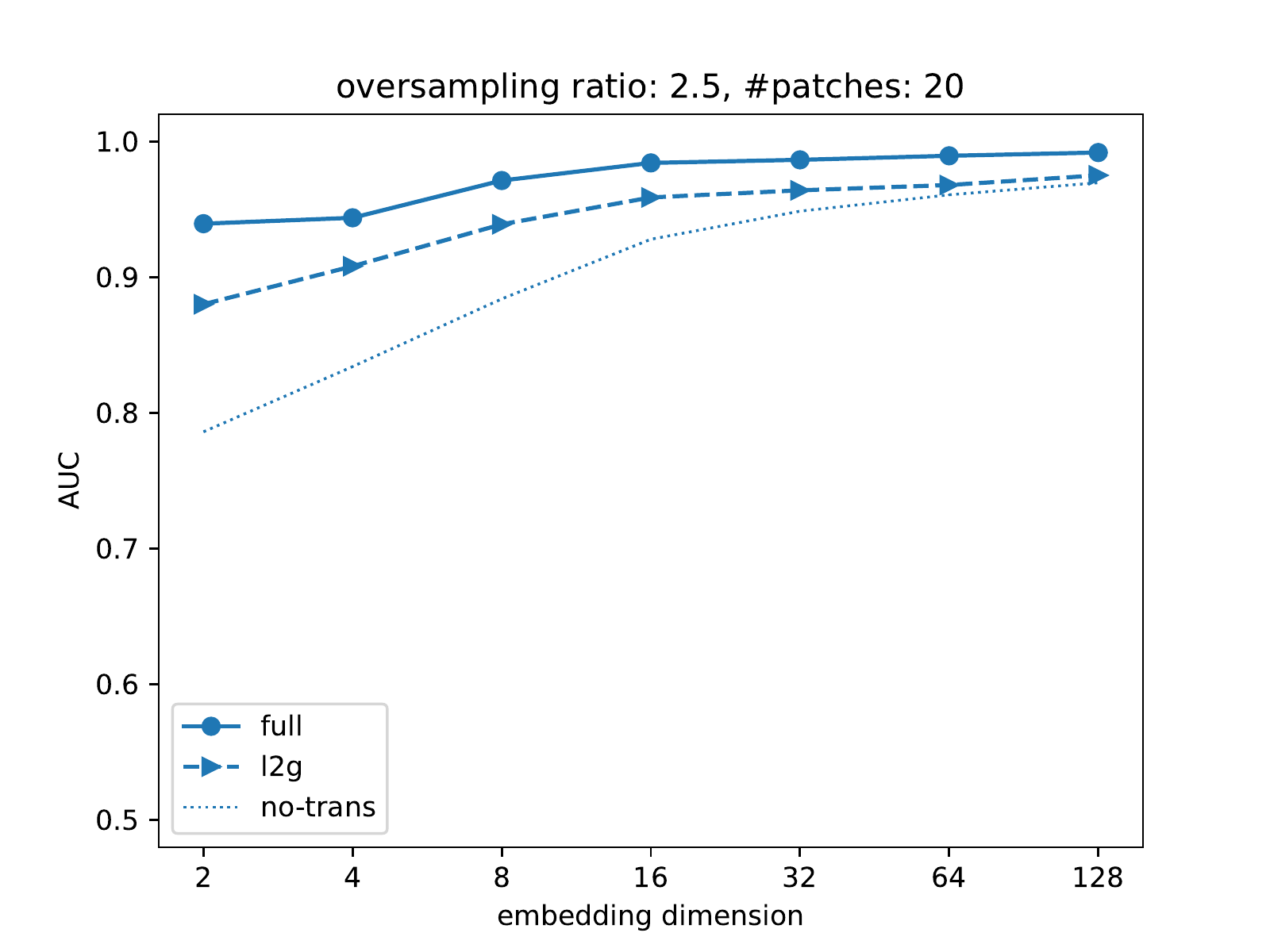}%
\end{minipage}}

\subfloat[\label{fig:VGAE_cl_scores:cora}]{%
\begin{minipage}{0.33\linewidth}
\includegraphics[width=\linewidth, clip, trim={0 0.2cm 0 0.8cm}]{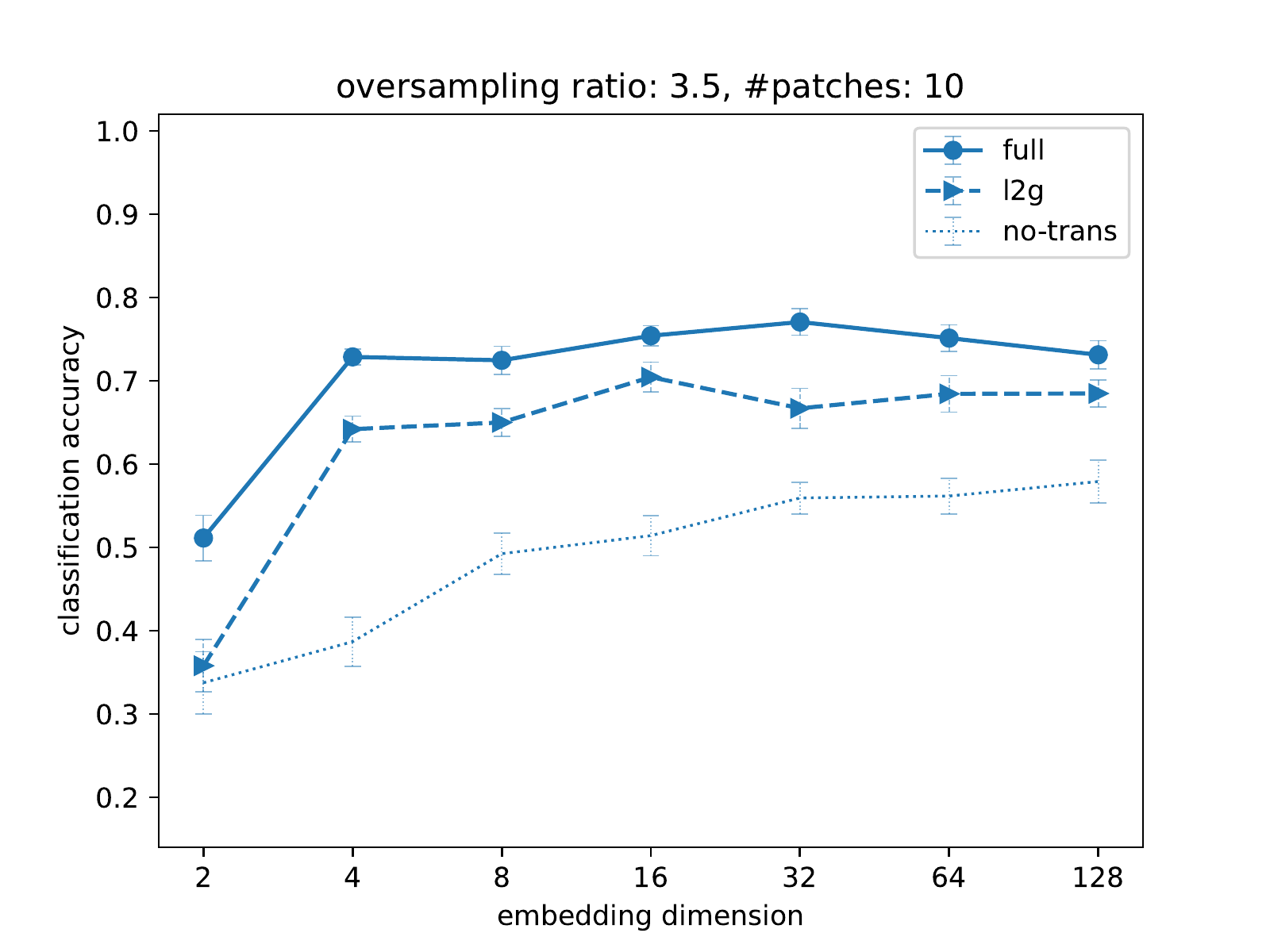}
\end{minipage}}\hfill
\subfloat[\label{fig:VGAE_cl_scores:amz_photo}]{%
\begin{minipage}{0.33\linewidth}
\includegraphics[width=\linewidth, clip, trim={0 0.2cm 0 0.8cm}]{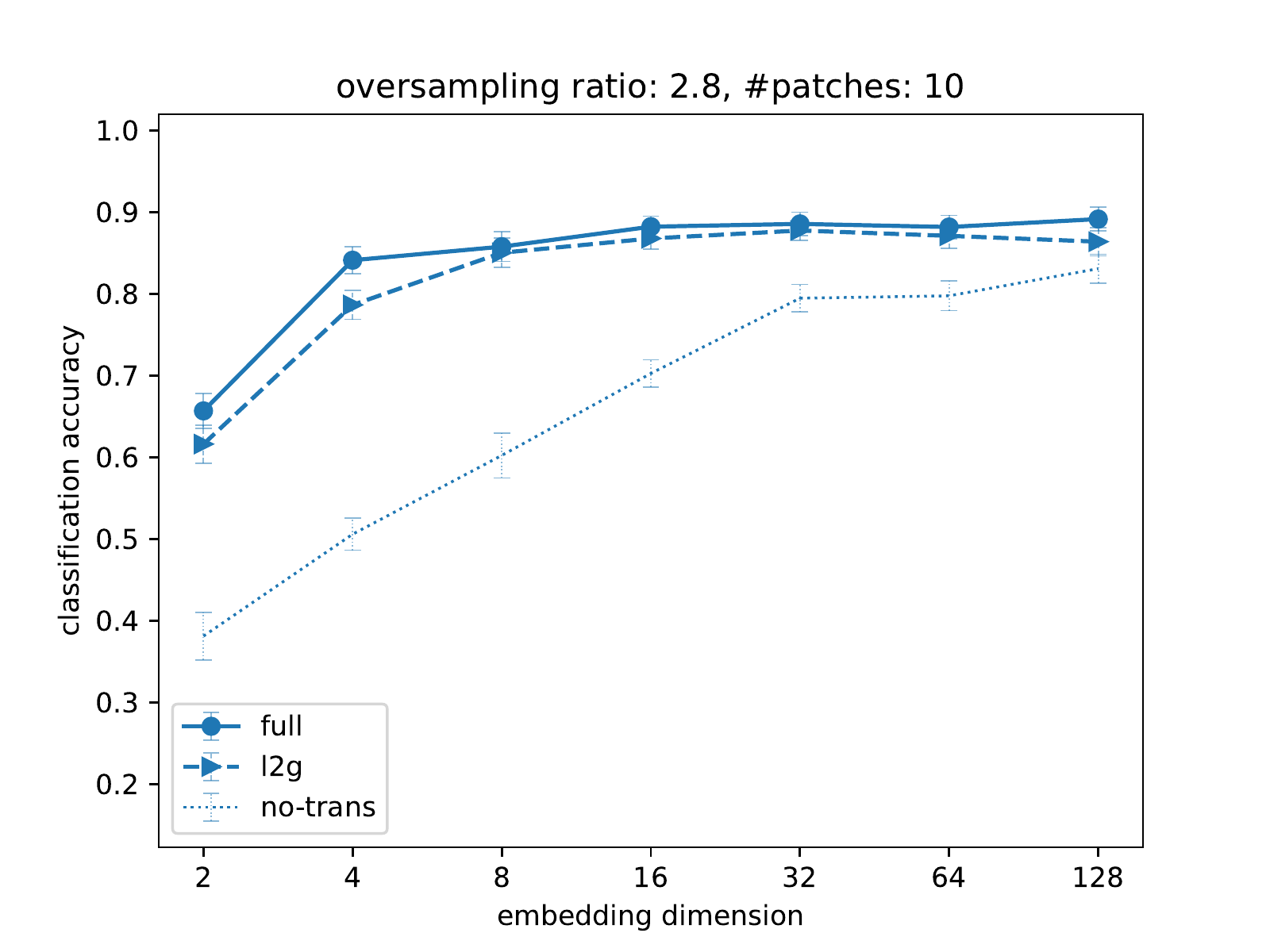}%
\end{minipage}}\hfill
\subfloat[\label{fig:VGAE_cl_scores:pubmed}]{%
\begin{minipage}{0.33\linewidth}
\includegraphics[width=\linewidth, clip, trim={0 0.2cm 0 0.8cm}]{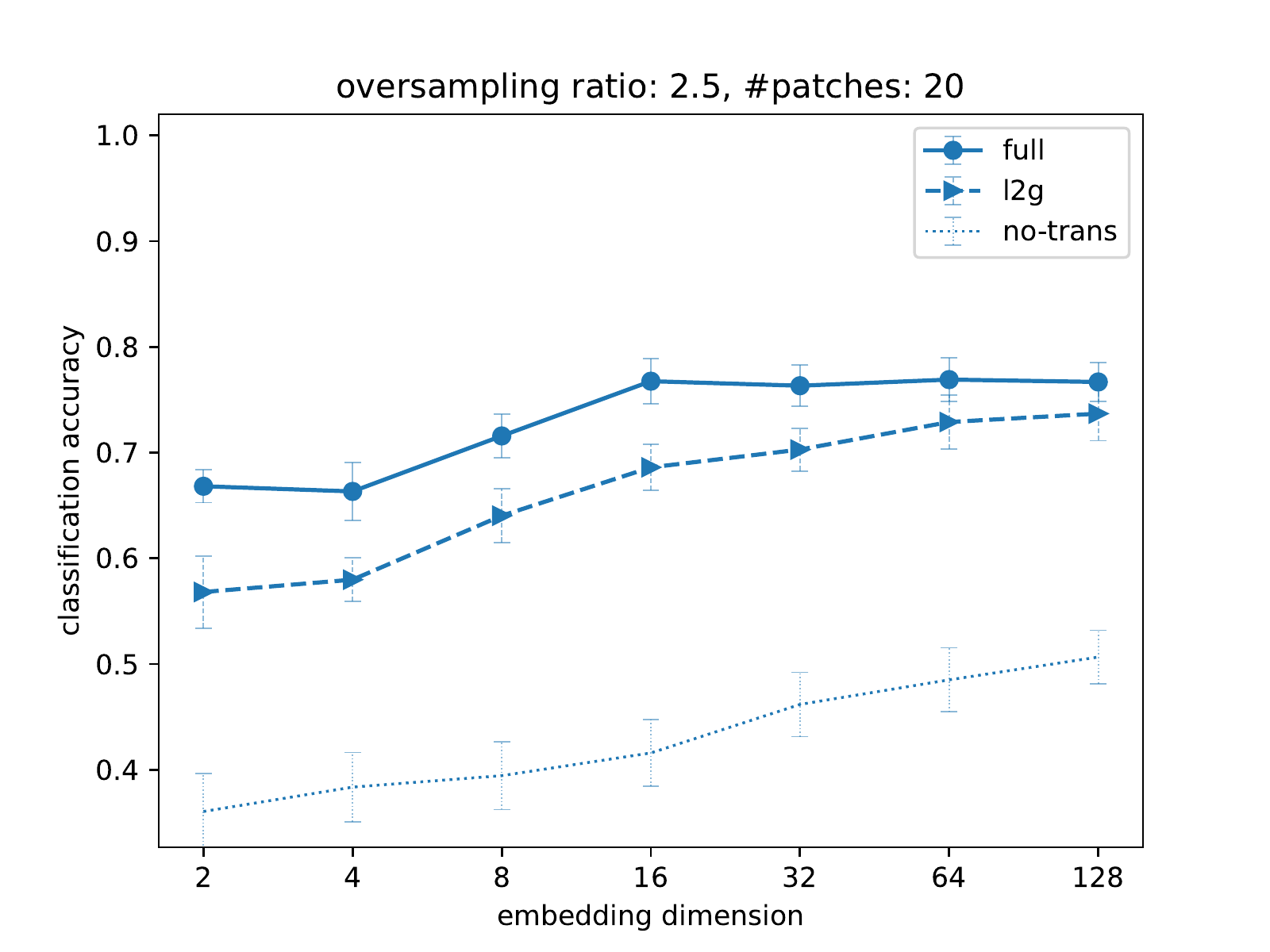}%
\end{minipage}}

\caption{VGAE embeddings: AUC network reconstruction score (top) and classification accuracy (bottom) as a function of embedding dimension using full data or stitched patch embeddings for \protect\subref{fig:VGAE_auc_scores:cora}, \protect\subref{fig:VGAE_cl_scores:cora} Cora, \protect\subref{fig:VGAE_auc_scores:amz_photo}, \protect\subref{fig:VGAE_cl_scores:amz_photo} Amazon photo, and \protect\subref{fig:VGAE_auc_scores:pubmed}, \protect\subref{fig:VGAE_cl_scores:pubmed} Pubmed. \label{fig:VGAE_scores}}
\end{figure*}

\begin{figure*}
\subfloat[\label{fig:DGI_auc_scores:cora}]{%
\begin{minipage}{0.33\linewidth}
\includegraphics[width=\linewidth, clip, trim={0 0.2cm 0 0.8cm}]{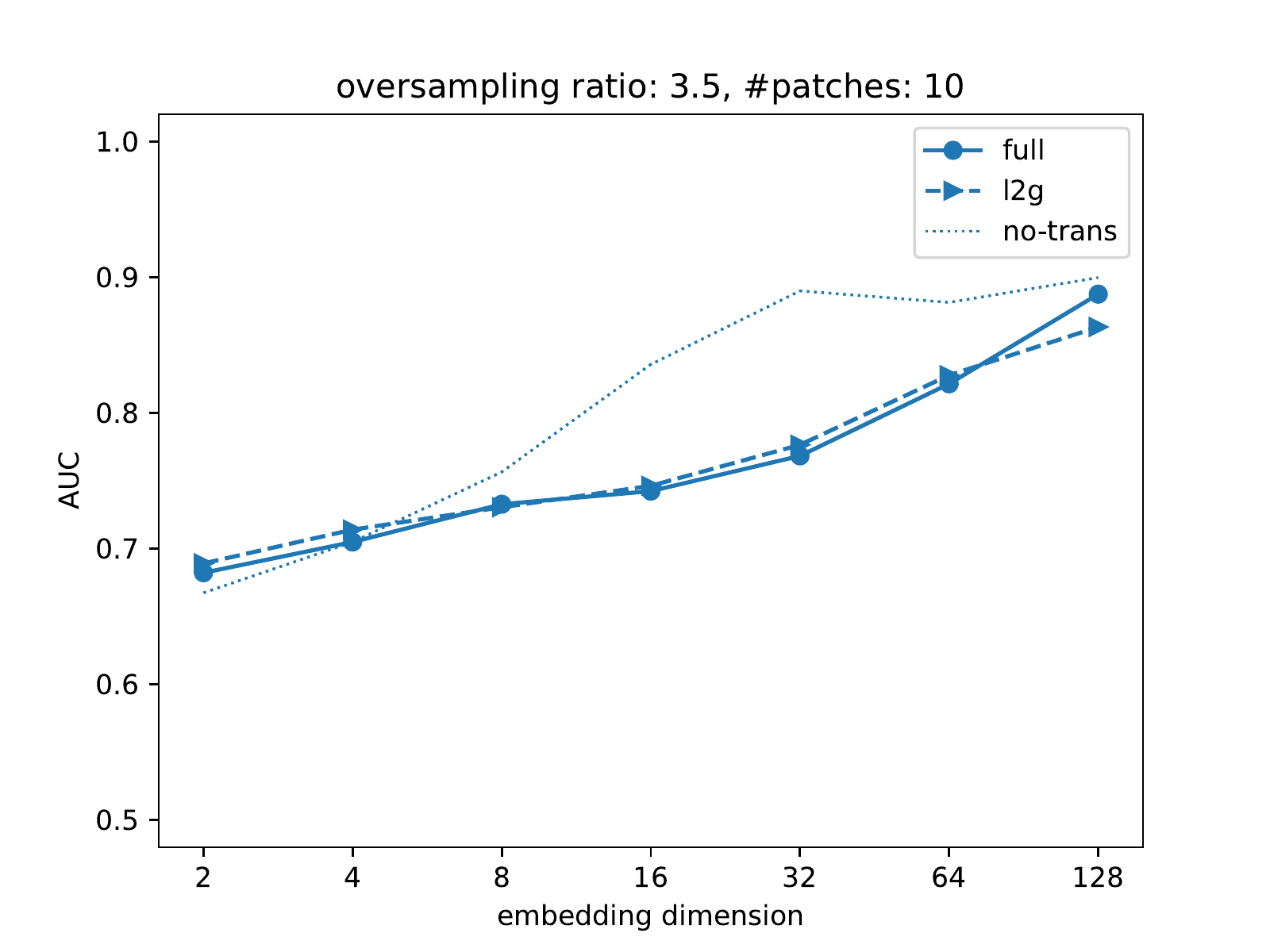}
\end{minipage}}\hfill
\subfloat[\label{fig:DGI_auc_scores:amz_photo}]{%
\begin{minipage}{0.33\linewidth}
\includegraphics[width=\linewidth, clip, trim={0 0.2cm 0 0.8cm}]{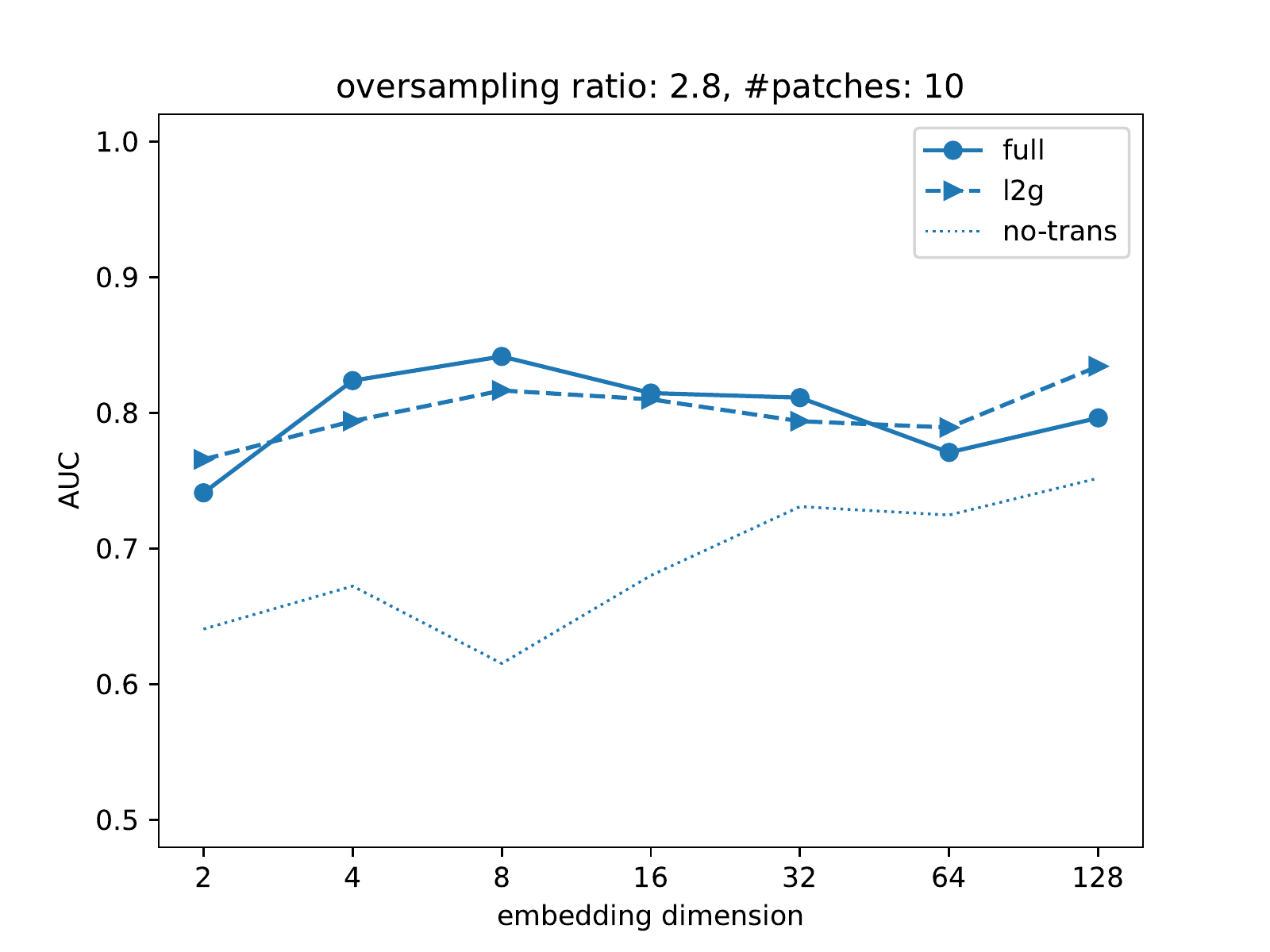}%
\end{minipage}}\hfill
\subfloat[\label{fig:DGI_auc_scores:pubmed}]{%
\begin{minipage}{0.33\linewidth}
\includegraphics[width=\linewidth, clip, trim={0 0.2cm 0 0.8cm}]{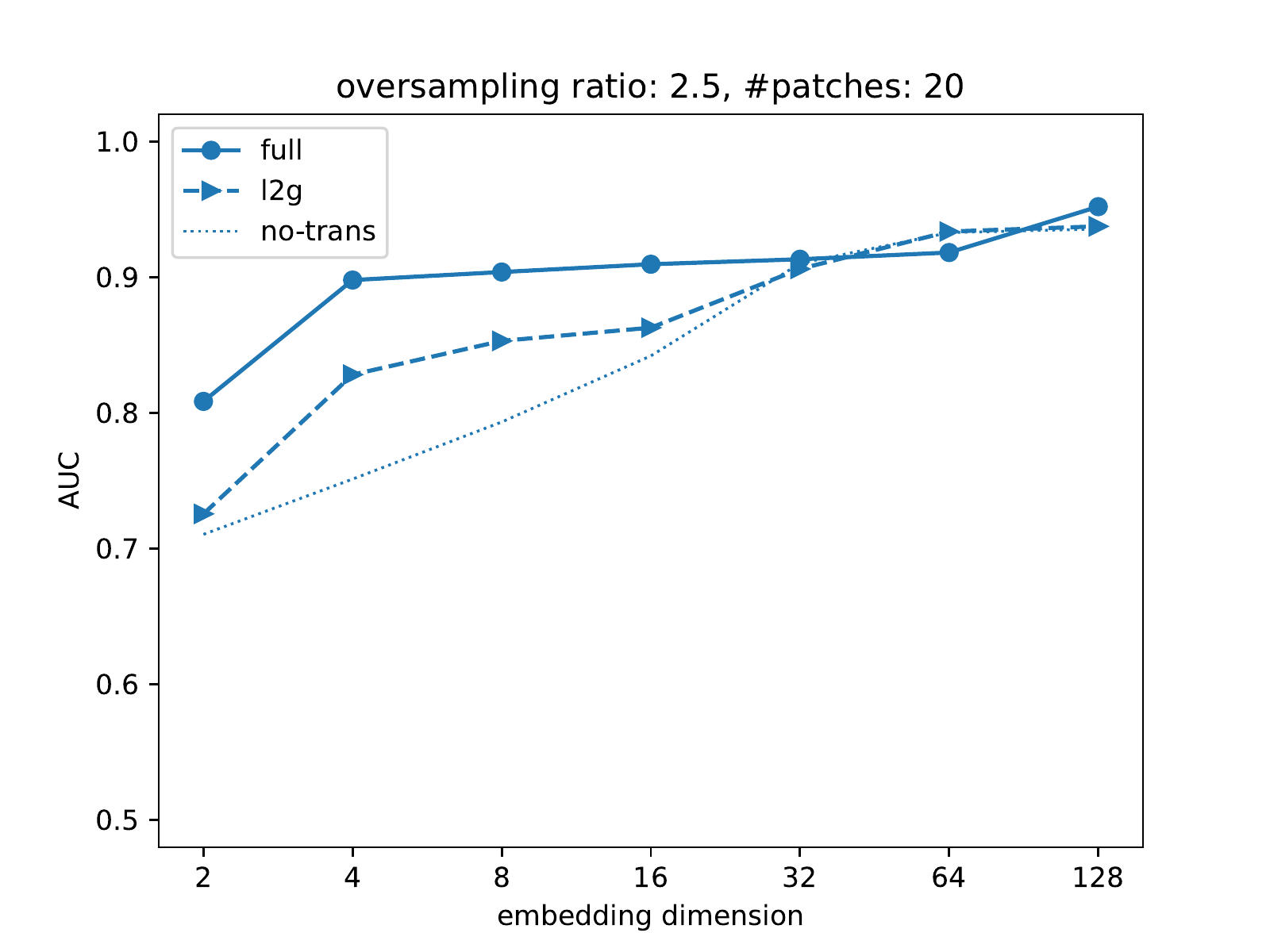}%
\end{minipage}}

\subfloat[\label{fig:DGI_cl_scores:cora}]{%
\begin{minipage}{0.33\linewidth}
\includegraphics[width=\linewidth, clip, trim={0 0.2cm 0 0.8cm}]{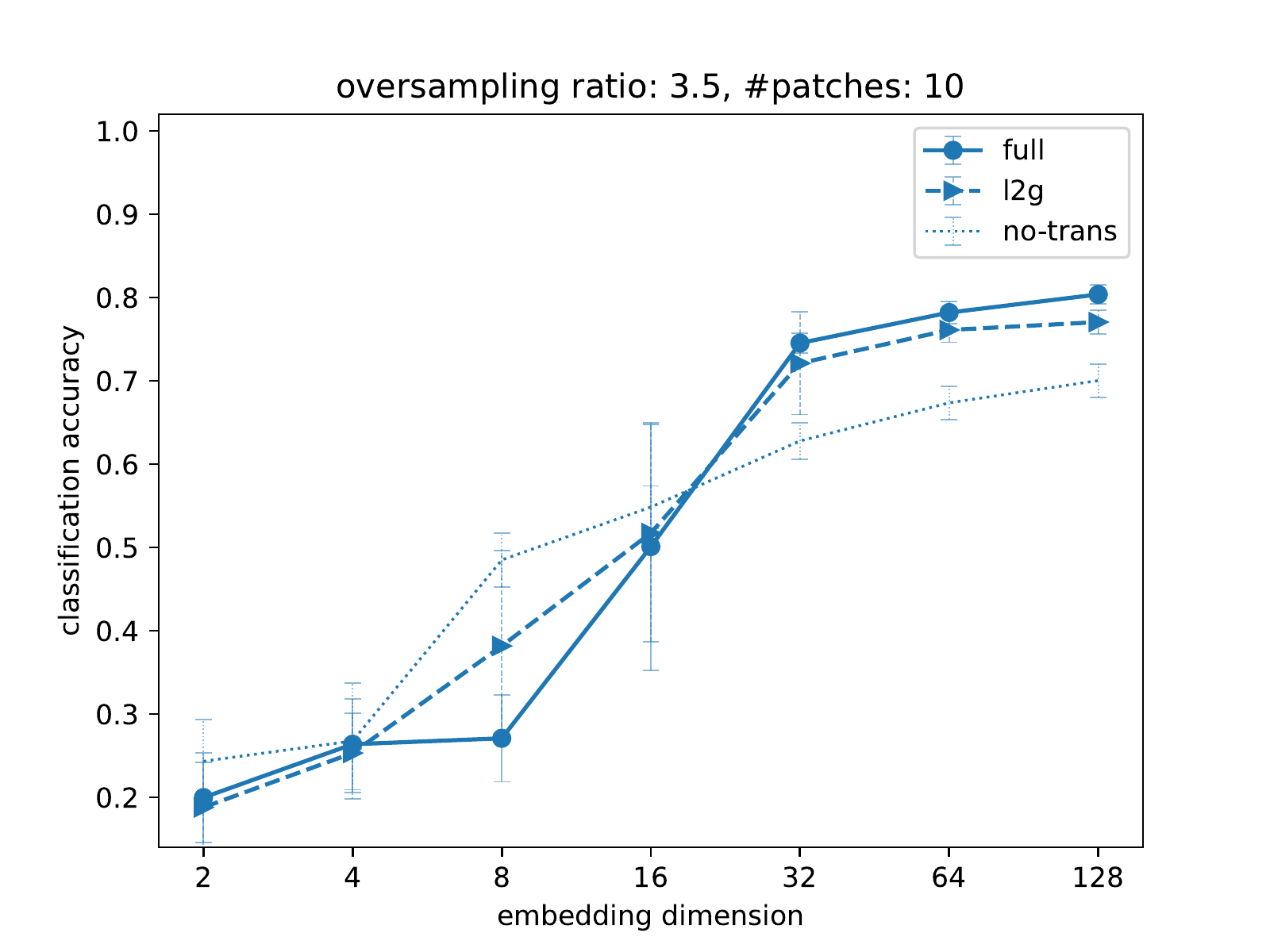}
\end{minipage}}\hfill
\subfloat[\label{fig:DGI_cl_scores:amz_photo}]{%
\begin{minipage}{0.33\linewidth}
\includegraphics[width=\linewidth, clip, trim={0 0.2cm 0 0.8cm}]{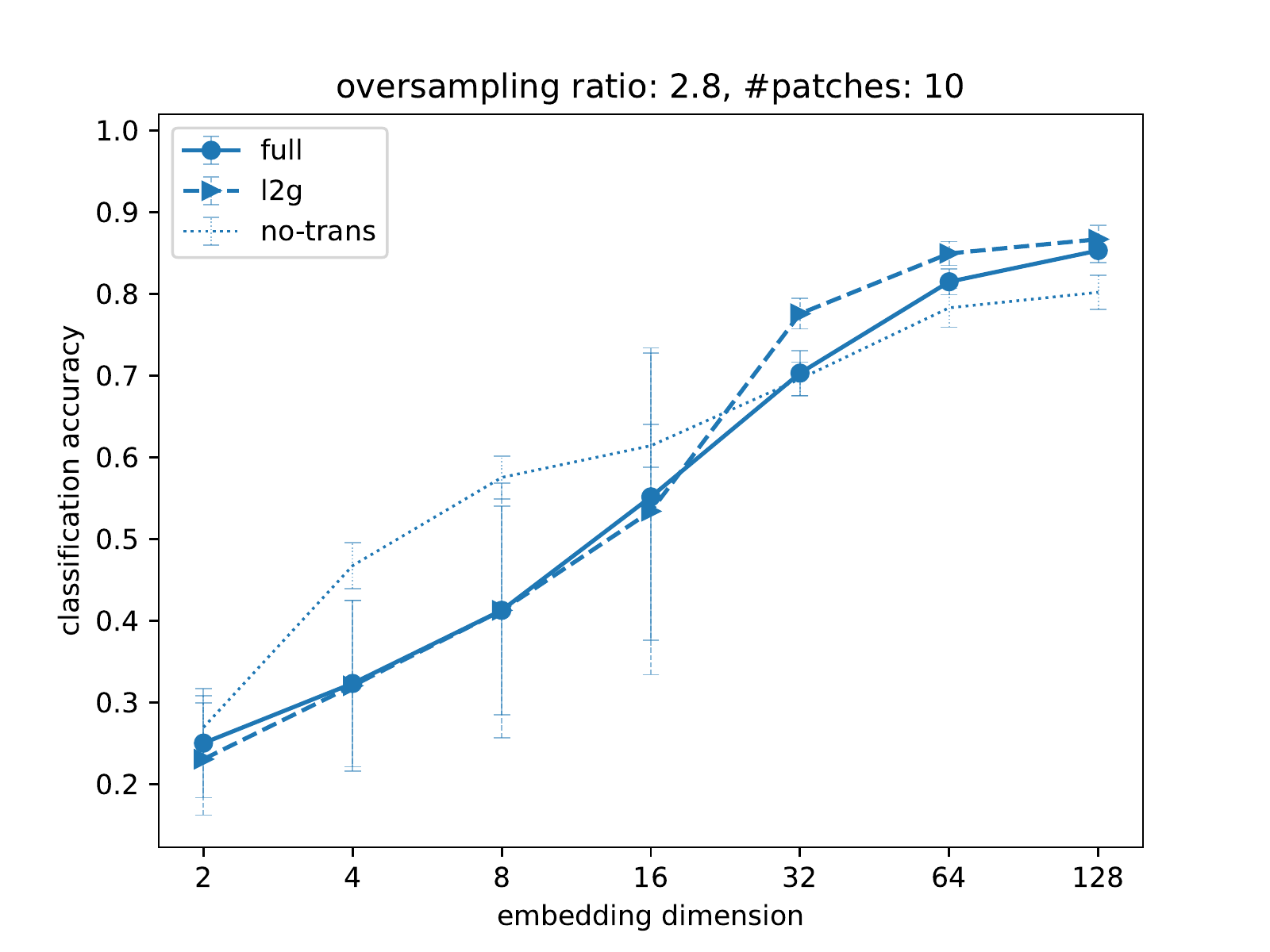}%
\end{minipage}}\hfill
\subfloat[\label{fig:DGI_cl_scores:pubmed}]{%
\begin{minipage}{0.33\linewidth}
\includegraphics[width=\linewidth, clip, trim={0 0.2cm 0 0.8cm}]{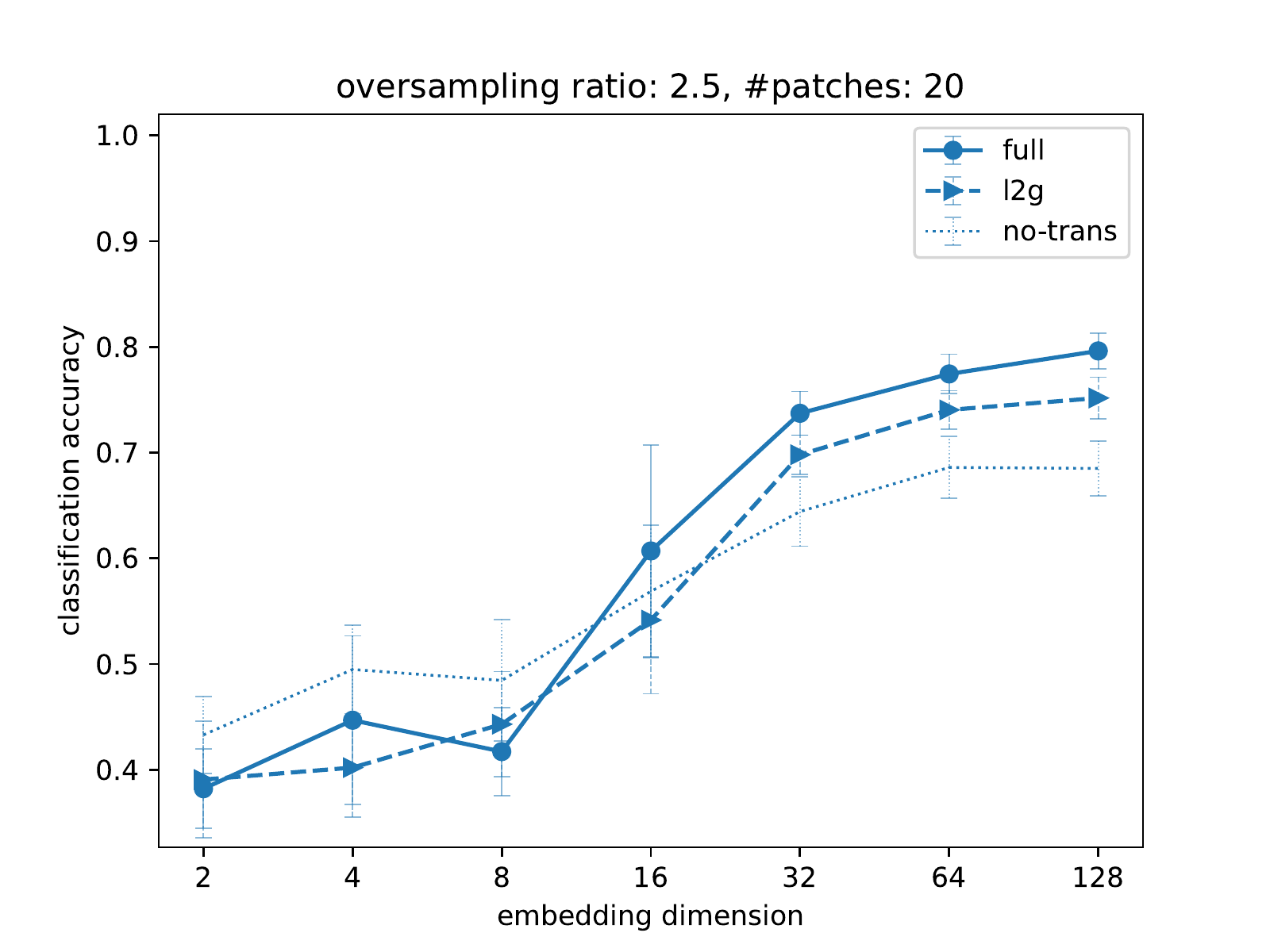}%
\end{minipage}}

\caption{DGI embeddings: AUC network reconstruction score (top) and classification accuracy (bottom) as a function of embedding dimension using full data or stitched patch embeddings for \protect\subref{fig:DGI_auc_scores:cora}, \protect\subref{fig:DGI_cl_scores:cora} Cora, \protect\subref{fig:DGI_auc_scores:amz_photo}, \protect\subref{fig:DGI_cl_scores:amz_photo} Amazon photo, and \protect\subref{fig:DGI_auc_scores:pubmed}, \protect\subref{fig:DGI_cl_scores:pubmed} Pubmed. \label{fig:DGI_scores}}
\end{figure*}

For the small-scale tests it is feasible to train the embedding models directly on the entire data set using full-batch gradient descent. We train the models using all edges in the largest connected component and compare three training scenarios
\begin{description}[leftmargin=\parindent]
\item[\bb full:] Model trained on the full data.
\item[\bb l2g:] Separate models trained on the subgraph induced by each patch and stitched using the {\ltg} algorithm.
\item[\bb no-trans:] Same training as l2g but node embeddings are obtained by taking the centroid over patch embeddings that contain the node without applying the alignment transformations. 
\end{description}

For Cora and Amazon photo, we split the network into 10 patches and sparsify the patch graph to a target mean degree $k=4$. For Pubmed, we split the network into 20 patches and sparsify the patch graph to a target mean degree of $k=5$.
We set the lower bound for the overlap to $l=256$ and upper bound to $u=1024$.

We show the results for VGAE embeddings in \cref{fig:VGAE_scores} and for DGI embeddings in \cref{fig:DGI_scores}. 
Overall, the gap between the results for `l2g` and `full` is small and essentially vanishes for higher embedding dimensions.
The aligned `l2g` embeddings typically outperform the unaligned `no-trans' baseline. Comparing the two embedding methods, we see that VGAE has better network reconstruction performance than DGI. This is not surprising, as network reconstruction corresponds to the training objective of VGAE. In terms of semi-supervised classification performance, the two methods are rather similar with no clear winner. However, DGI requires higher embedding dimensions for good performance. 

\subsubsection{Large-scale tests}

For the large-scale MAG240m network we use two iterations of FENNEL \citep{Tsourakakis2014} to divide the network into \num{10000} clusters. We set the lower bound for the overlap to $l=512$ and the upper bound to $u=2048$ and sparsify the patch graph to have a mean degree $\langle k \rangle=20$.

\begin{table}[ht]
\centering
\begin{tabular}{r|S[table-number-alignment=right,table-format=1.2]S[table-number-alignment=right,table-format=1.2]}
& {l2g} & {no-trans}\\
\hline
reconstruction (AUC) & 0.96 & 0.70\\
classification (Acc.) & 0.54 & 0.42
\end{tabular}
\caption{Network reconstruction (AUC) and classification accuracy for ${d=128}$ VGAE embeddings of MAG240m. \label{tab:mag240m_vgae}}
\end{table}

\begin{table}[ht]
\centering
\begin{tabular}{r|S[table-number-alignment=right,table-format=1.2]S[table-number-alignment=right,table-format=1.2]}
& {l2g} & {no-trans}\\
\hline
reconstruction (AUC) & 0.91 & 0.69\\
classification (Acc.) & 0.53 & 0.45
\end{tabular}
\caption{Network reconstruction (AUC) and classification accuracy for ${d=128}$ DGI embeddings of MAG240m. \label{tab:mag240m_dgi}}
\end{table}

We show the results for VGAE in \cref{tab:mag240m_vgae} and for DGI in \cref{tab:mag240m_dgi}. 
For the semi-surpervised classification results we use a grid search over the parameters of the MLP as in \cite{hu2021ogblsc}, with hidden dimension in $\{128, 256, 512, 1024\}$, number of layers in $\{2, 3, 4\}$, and dropout probability in $\{0, 0.25, 0.5\}$. 
We fix the batch size to \num{100000} and use the Adam optimizer \citep{Kingma2015} with learning rates in $\{0.01, 0.001, 0.0001\}$. We use early stopping based on the validation loss with patience of 20 epochs and maximum number of epochs set to 1000. 
We use batch normalisation \citep{ioffe2015} after each layer together with a ReLU non-linearity. 
We search over the hyperparameters separately for each embedding and report results for the model with the highest validation accuracy. 
However, the best parameter setting is consistent across all embeddings reported in \cref{tab:mag240m_dgi,tab:mag240m_vgae} with the highest validation accuracy achieved by a 4-layer MLP with hidden dimension 1024 trained with learning rate set to 0.0001. 
There is a clear improvement when comparing the results for {\ltg} (l2g) with the no transformation baseline (no-trans), and performance results for network reconstruction are high for both embedding methods. While the classification accuracy we obtain with {\ltg} in \cref{tab:mag240m_dgi} is on par with that of MLP in \cite{hu2021ogblsc}, we note that the classification accuracy results are not directly comparable. All approaches in \cite{hu2021ogblsc} differ from the {\ltg} pipeline in at least one of two fundamental ways: (1) the training step is supervised and (2) there is no dimensionality reduction on a feature level.

\section{Application to anomaly detection}
\label{sec:anomaly}

As an application of our {\ltg} algorithm, we consider the task of anomaly detection in a temporal cyber-security data set. In particular, we consider the netflow part of the Los Alamos National Laboratory Unified Host and Network Data Set (LANL-netflow, \cite{Turcotte2018}). LANL-netflow consists of 89 days of processed network flow records (see \cite{Turcotte2018} for details). For our experiments we aggregate the network flows per day, restricted to the TCP protocol. We treat the network as directed and unweighted, only considering the presence or absence of a directed connection. For the embeddings, we consider each day as a patch and treat the network as bipartite, computing separate source and destination embeddings for each node using the SVD-based method of \cite{Dhillon2001}.

\subsection{Patch embeddings}
In particular, let $\mat{A}^{(t)}$ be the adjacency matrix for the network at day $t$, where $A^{(t)}_{ij}=1$ if there was a directed connection between source node $i$ and destination node $j$ during day $t$ and $A^{(t)}_{ij}=0$ otherwise. We consider the normalized adjacency matrix $\mat{A}_n^{(t)} = {\mat{D}_s^{(t)}}^{-\frac{1}{2}} \mat{A}^{(t)} {\mat{D}_d^{(t)}}^{-\frac{1}{2}}$, where $\mat{D}_s^{(t)}$ and $\mat{D}_d^{(t)}$ are the diagonal source and destination degree matrices respectively, i.e., ${\mat{D}_s^{(t)}}_{ii} = \sum_j \mat{A}_{ij}^{(t)}$ and ${\mat{D}_d^{(t)}}_{jj} = \sum_i \mat{A}^{(t)}_{ij}$. We define the (unaligned) source embedding $\mat{X}^{(t)}$ and destination embedding $\mat{Y}^{(t)}$ based on the singular value decomposition of $\mat{A}_n^{(t)}$ such that $\mat{X}^{(t)} = {\mat{D}_s^{(t)}}^{-\frac{1}{2}} \mat{U}^{(t)}$ and $\mat{Y}^{(t)} = {\mat{D}_d^{(t)}}^{-\frac{1}{2}} \mat{V}^{(t)}$ where $\mat{U}^{(t)}$ is the matrix with columns given by the $d$ leading left singular vectors and $\mat{V}^{(t)}$ the matrix with columns given by the $d$ leading right singular vectors of $\mat{A}_n^{(t)}$ in the space orthogonal to the trivial solution, i.e., $\vec{1}{\mat{D}_s^{(t)}}^{\frac{1}{2}} \mat{U}^{(t)} = \vec{0}$ and $\vec{1}{\mat{D}_d^{(t)}}^{\frac{1}{2}} \mat{V}^{(t)} = \vec{0}$.

\subsection{Reference embedding}

We use the {\ltg} algorithm to construct a reference embedding to use as a baseline for identifying anomalous behavior. As mentioned above, we treat each day of observations as a patch. Computers are largely persistent on the network from one day to the next, such that we have sufficiently many overlapping nodes between different patches to reliably estimate the relative transformations. We define the patch graph $G_p(\set{P}, E_p)$ based on the delay between patches, such that $\{P_s, P_t\} \in E_p \iff \abs{s-t} \in \{1, 7, 14, 21, 28, 35, 42, 49\}$, i.e., we compute relative transformations between patches that are one day and multiples of one week apart. In preliminary experiments, we found that including long-range connections in the patch graph is crucial to obtaining good results in this application. We separately apply the {\ltg} algorithm to obtain a reference source embedding $\bar{\mat{X}}$ and a reference destination embedding $\bar{\mat{Y}}$ which are defined as the centroid over the  aligned patch source embeddings $\{\bar{\mat{X}}^{(t)}\}$ and destination embeddings $\{\bar{\mat{Y}}^{(t)}\}$ (see \cref{eq:mean_embedding}). Note that we only compute a single reference embedding based on all patches. In practical applications, one would apply the alignment in a streaming fashion, such that the reference embedding is only based on past data and updated as new data becomes available. However, given the limited snapshot of data available as part of the LANL-netflow data set, applying such a streaming algorithm is problematic.

\subsection{Anomaly score}

We define the raw source anomaly score $\Delta_{s_i}^{(t)}= \norm{\bar{\mat{X}}_i - \bar{\mat{X}}_i^{(t)}}_2$ for source node $i$ at time $t$, and raw destination anomaly score $\Delta_{d_j}^{(t)} = \norm{\bar{\mat{Y}}_j -\bar{\mat{Y}}_j^{(t)}}_2$ for destination node $j$ at time $t$ based on the Euclidean distance between the  reference embedding and the aligned patch embeddings. The raw scores are not comparable across nodes, as the behavior and hence the embedding is inherently more stable for some nodes than others.

We use a z-score transformation to standardize the scores and make them comparable between nodes to enable easier identification of anomalous behavior. We define the standardized anomaly scores as
\begin{equation}
    \hat{\Delta}_{s_i}^{(t)} = \frac{\Delta_{s_i}^{(t)} - \mu_{-t}(\Delta_{s_i})}{\sigma_{-t}(\Delta_{s_i})}, 
    \qquad
    \hat{\Delta}_{d_j}^{(t)} = \frac{\Delta_{d_j}^{(t)} - \mu_{-t}(\Delta_{d_j})}{\sigma_{-t}(\Delta_{d_j})},
    \label{eq:z-score}
\end{equation}
where we use $\mu_{-t}$ and $\sigma_{-t}$ to denote the mean and standard deviation estimated based on all data points except the data point at time $t$. Leaving out the current point when estimating the mean and standard deviation has the effect of boosting the scores for outliers and thus leads to a clearer signal.

\subsection{Results}

\begin{figure}
    \centering
    \subfloat[source\label{sfig:reference:source}]{\includegraphics[width=0.48\linewidth]{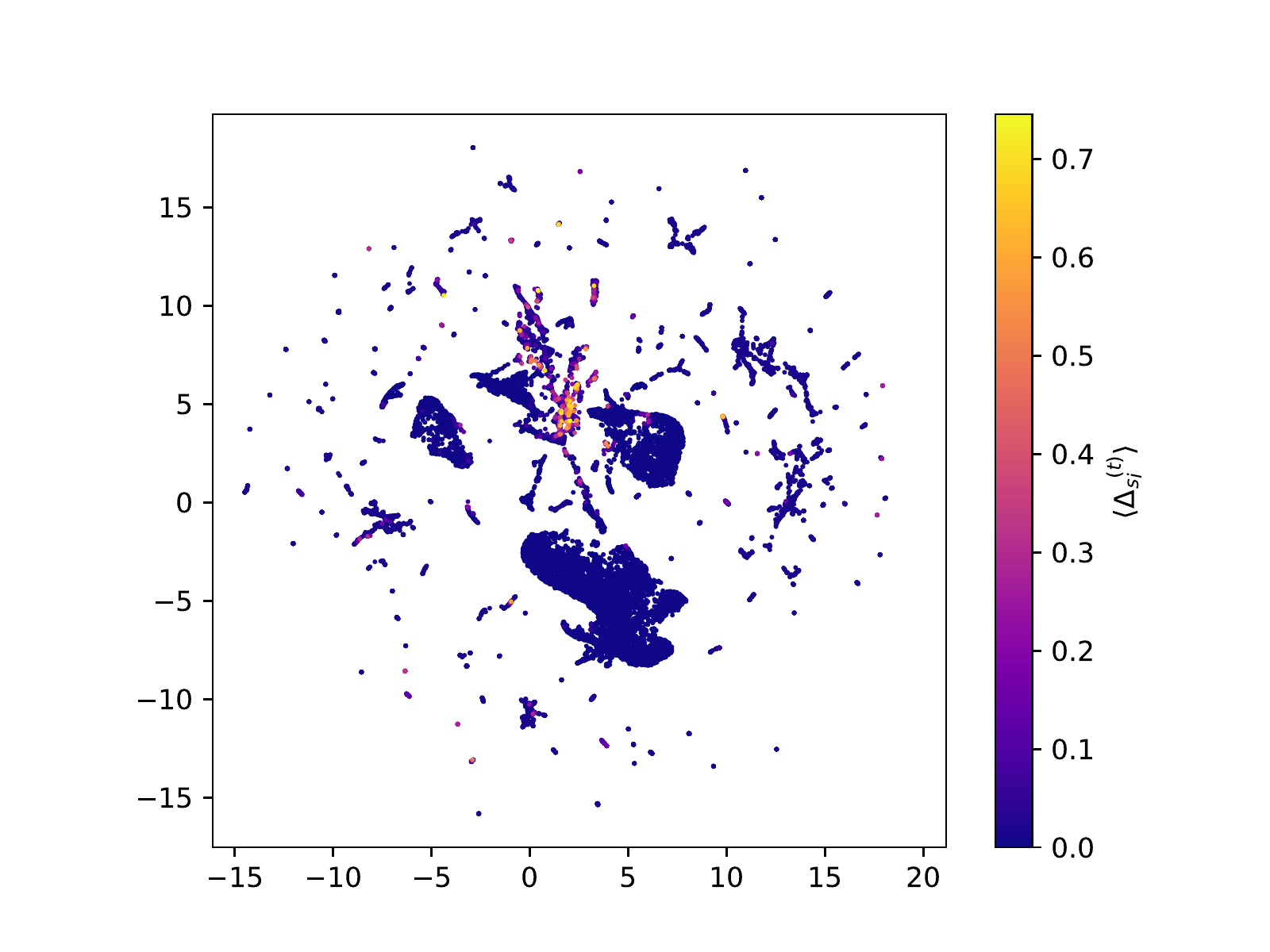}}\hfill
    \subfloat[destination\label{sfig:reference:dest}]{%
    \includegraphics[width=0.48\linewidth]{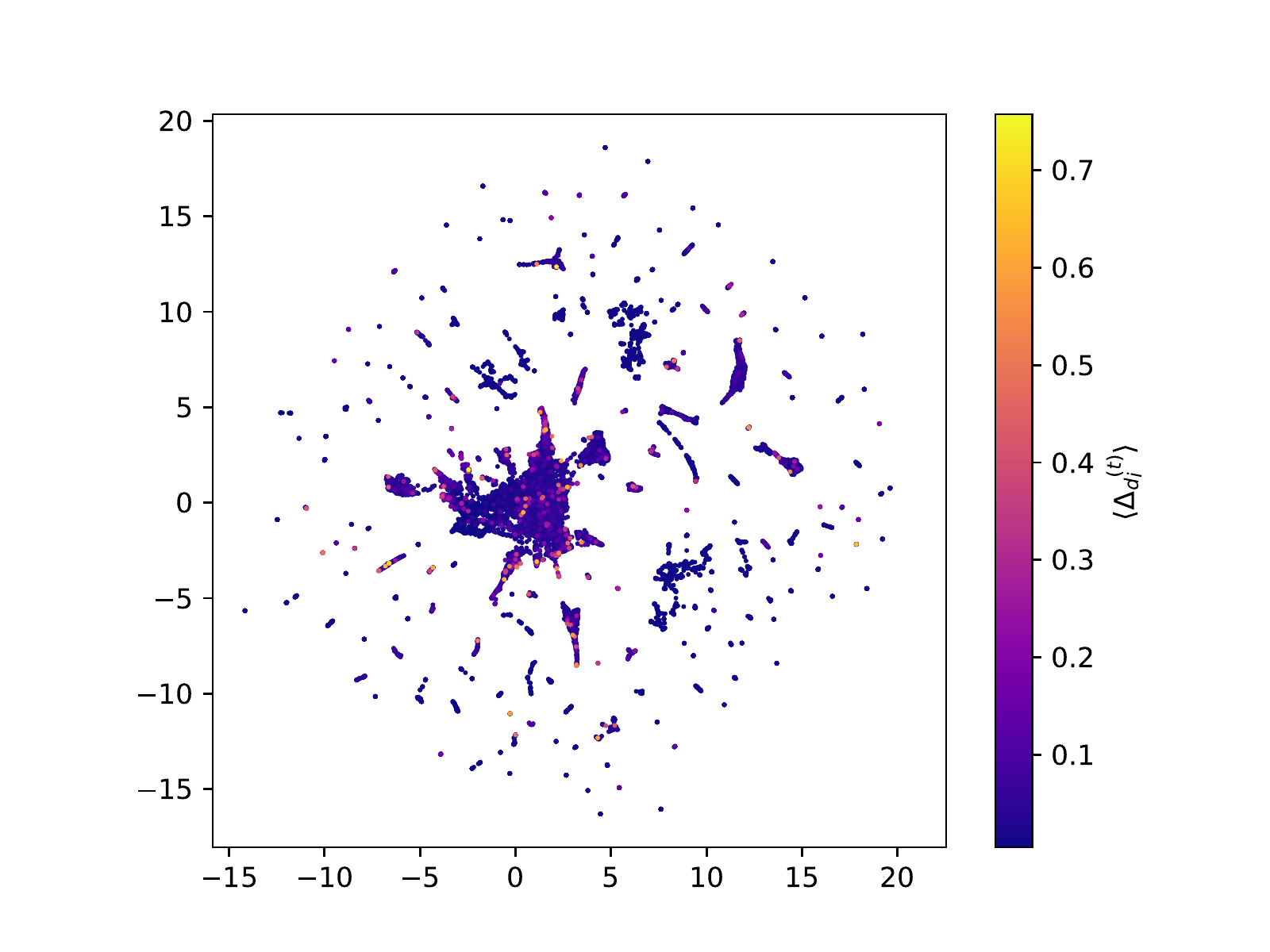}}
    \caption{UMAP visualisation of average alignment errors for \protect\subref{sfig:reference:source} source embeddings and \protect\subref{sfig:reference:dest} destination embeddings. Only nodes with at least 21 days of observations are shown in the visualisation.
    \label{fig:reference}}
\end{figure}

\begin{figure*}
    \renewcommand{\footnoterule}{}
    \begin{minipage}{\linewidth}
    \centering
    \subfloat[source]{
    \begin{minipage}{0.48\linewidth}
    \includegraphics[width=\linewidth]{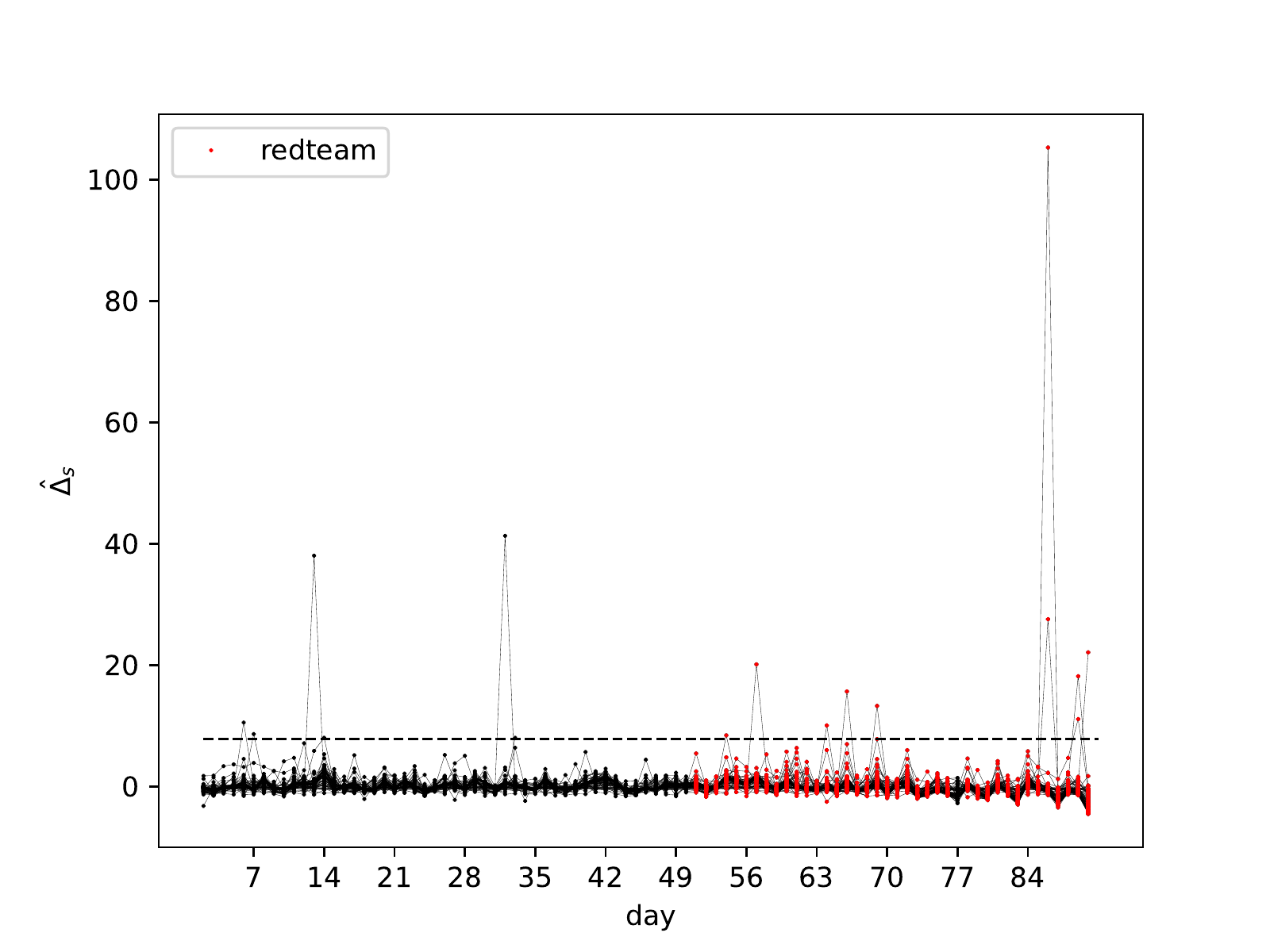}\\
    \includegraphics[width=\linewidth]{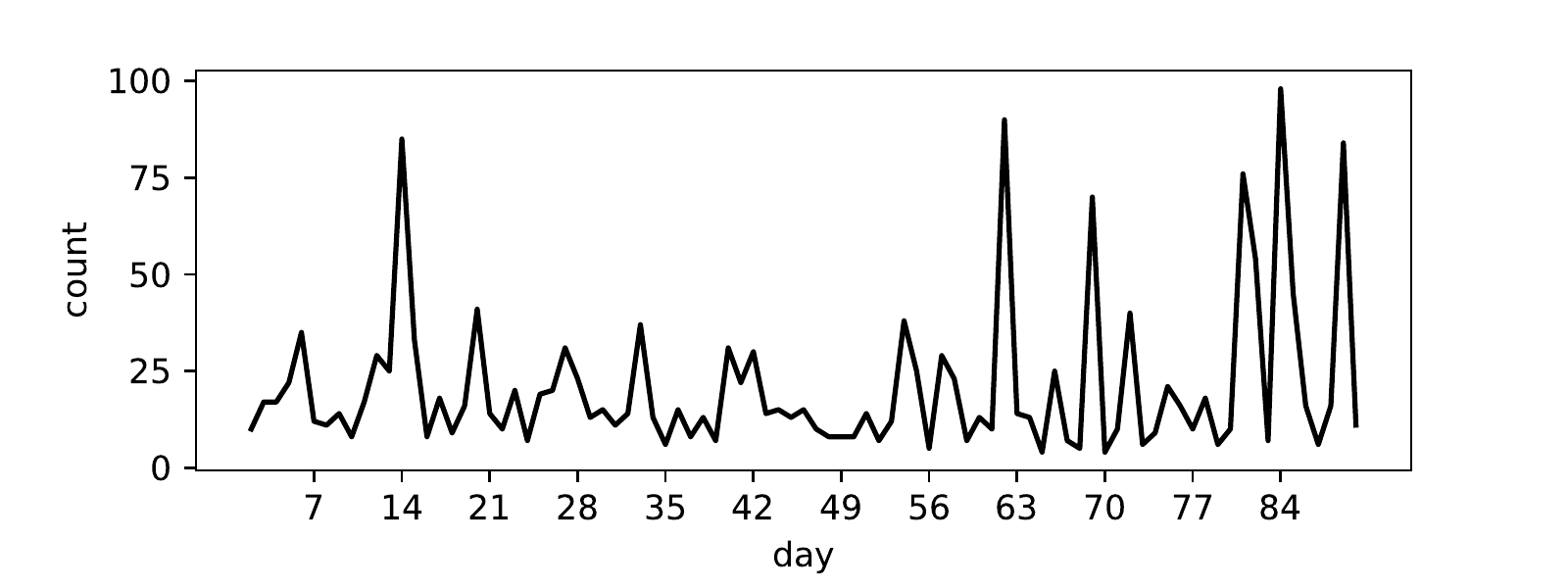}
    \end{minipage}}\hfill
    \subfloat[destination]{
    \begin{minipage}{0.48\linewidth}
    \includegraphics[width=\linewidth]{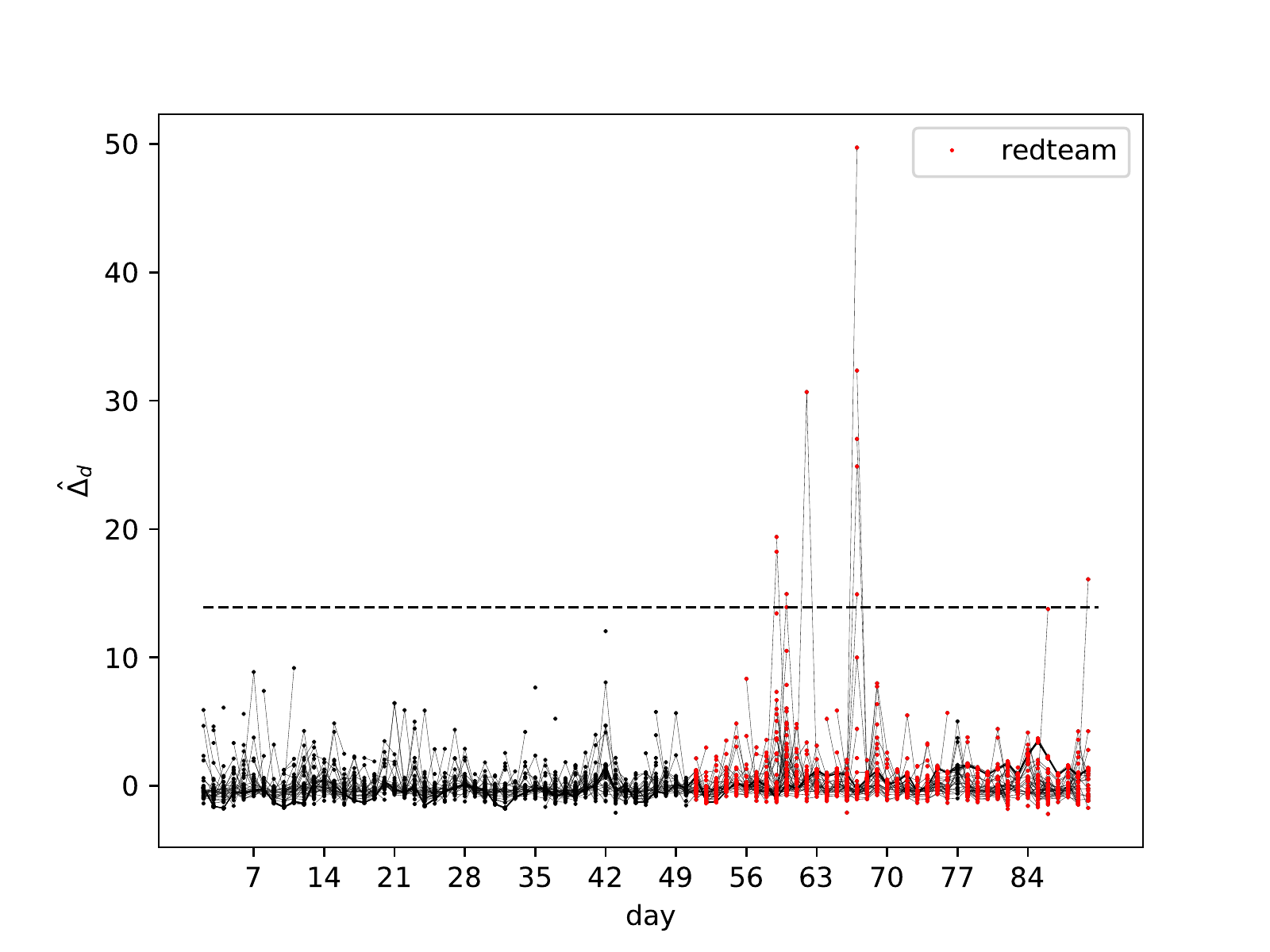}
    \includegraphics[width=\linewidth]{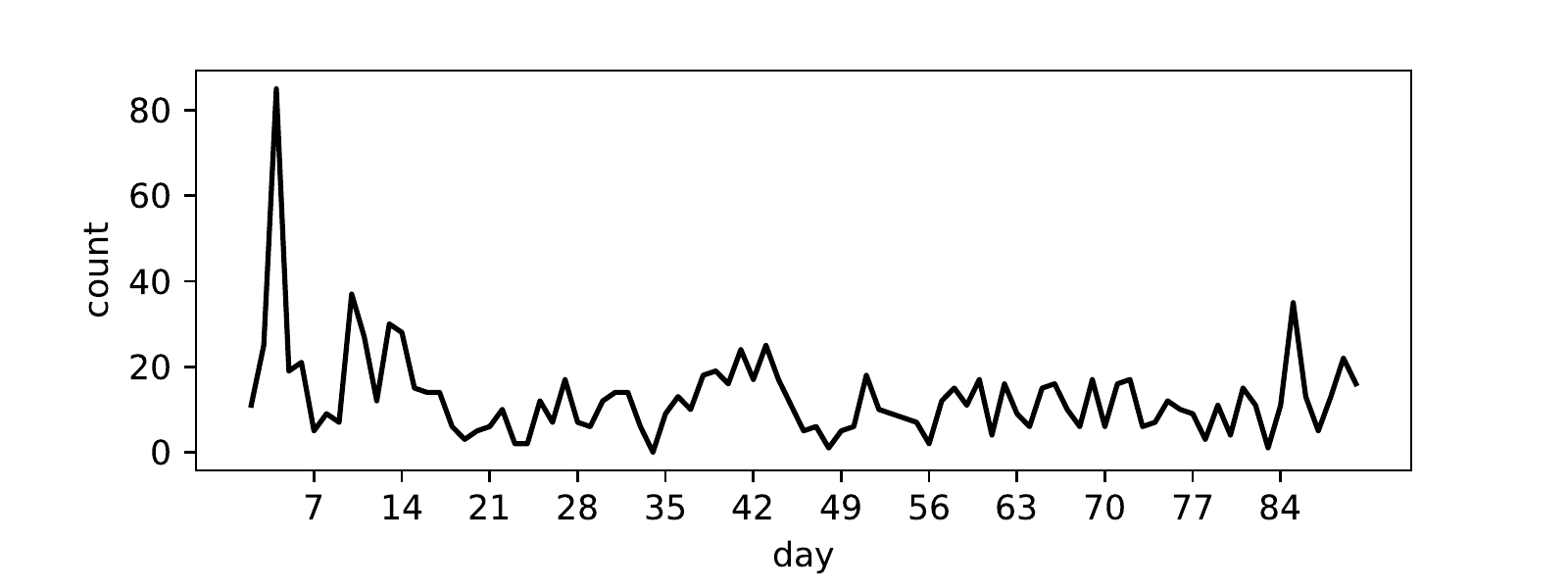}
    \end{minipage}
    }
    \caption{Standardized anomaly scores for known red-team targets (top) and outlier count\protect\footnote{Outliers are defined as nodes with scores in the 0.999 quantile of the score distribution across days and nodes.} (bottom)  for (a) source embeddings and (b) destination embeddings. The dashed line indicates the 0.999 quantile of the anomaly score distribution over all days and nodes with at least 21 days of observations. Days with observed redteam activity are highlighted in red.}
    \label{fig:z-score}
    \end{minipage}
\end{figure*}

We visualize the reference embeddings $\bar{\mat{X}}$ and $\bar{\mat{Y}}$ together with the raw anomaly scores in \cref{fig:reference} using UMAP \citep{mcinnes2020umap}. For most nodes, the embeddings are very stable across time with only small average error between aligned patch embeddings and the reference embedding. However, some nodes exhibit much noisier behavior with large average error. This difference in behavior is normalized out by the z-score transformation of \cref{eq:z-score} such that it is possible to define a common threshold for anomalous behavior. Another source of potential issues are nodes with insufficient observations to reliably estimate normal behavior. While we use all nodes to compute the embeddings for each day and the reference embedding, we restrict the results in \cref{fig:reference,fig:z-score} to nodes with at least 21 days of observations.

\Cref{fig:z-score} shows the standardized anomaly scores for the nodes with known red-team activity. We see a clear increase in the number of outliers for the affected nodes during the attack period, especially for the destination embedding. This suggests that red-team activity may have been responsible for some unusual network activity. However, it should be noted that there is no clear separation between red-team activity and other sources of unusual activity, as we observe outliers of similar magnitude for other nodes throughout the data. This suggests that network flows may provide an additional useful source of signal for identifying suspicious behavior. Combining network flows with node-level information from the process logs using, for e.g., a GCN-based embedding approach such as VGAE or DGI, could be interesting. However, extracting useful features from the process data is not straightforward, and we leave this task as future work.

\section{Conclusion}
\label{sec:conc}

In this work, we introduced a framework that can significantly improve the computational scalability of generic graph embedding methods, rendering them scalable to real-world applications that involve massive graphs, potentially with millions or even billions of nodes.
 At the heart of our pipeline is the {\ltg} algorithm, a divide-and-conquer approach that first decomposes the input graph into overlapping clusters (using one's method of choice), computes entirely-local embeddings via the  preferred embedding method for each resulting cluster (exclusively using information available at the nodes within the cluster), and finally stitches the resulting local embeddings into a globally consistent embedding, using established  machinery from the \textit{group synchronization} literature.
Our results on small-scale data sets achieve comparable accuracy on graph reconstruction and semi-supervised clustering as globally trained embeddings. We have also demonstrated that {\ltg} achieves a good trade-off between scalability and accuracy 
on large-scale data sets using a variety of embedding techniques and downstream tasks. Our results also suggest that the application of {\ltg} to the task of anomaly detection is 
fruitful with the potential for several future work directions.

More specifically, a first direction is to further increase the size of node and edge sets by another order of magnitude, and consider web-scale graphs with billions of nodes, which many existing algorithms will not be able to handle and for which distributed training is crucial. One can use our hierarchical implementation of {\ltg} (\cref{sec:hierchical}) to further increase the scalability of our approach, and explore the trade-off between scale and embedding quality within this hierarchical framework. A related direction is to explore the interplay of our pipeline with different clustering algorithms, and in particular hierarchical clustering, and assess how the patch graph construction mechanism affects the overall performance. 

A second direction is to further demonstrate particular benefits of locality and asynchronous parameter learning. These have clear advantages for privacy preserving and federated learning setups. It would also be particularly interesting to identify further settings in which this {\ltg} approach can outperform global methods.  The intuition 
and hope in
this direction stems from the fact that asynchronous locality can be construed as a regularizer (much like sub-sampling, and similar to dropout) and could potentially lead to better generalization and alleviate the oversmoothing issues of deep GCNs, as observed in \cite{Chiang2019}. 

Finally, the application of {\ltg} to anomaly detection is promising and shows that network flows could provide useful additional signal for identifying network intrusions, which is a particularly challenging task given the noisy nature of the data and the rarity of malicious events. 
It would be interesting to apply the methodology in \cref{sec:anomaly} to further data sets \citep{Highnam2021,optc}, and to incorporate node-level attributes into the embedding techniques. 
Generating meaningful node-level features for anomaly detection (e.g., based on device logs) and incorporating a spatio-temporal dimension into the analysis also warrant further investigation.

\backmatter

\bmhead{Acknowledgments}

The authors would like to acknowledge support from the Defence and Security Programme at The Alan Turing Institute, funded by the Government Communications Headquarters (GCHQ).

\bibliography{local2global}

\end{document}